%% file: egpaper_final.tex
\documentclass[10pt,twocolumn,letterpaper]{article}

\usepackage{iccv}
\usepackage{times}
\usepackage{epsfig}
\usepackage{graphicx}
\usepackage{amsmath}
\usepackage{amssymb}
\usepackage{graphicx}
\usepackage{amsmath}
\usepackage{amssymb}
\usepackage{booktabs}
\usepackage{bm}
\usepackage{times}
\usepackage{epsfig}
\usepackage{mathrsfs}
\usepackage{blindtext}
\usepackage[english]{babel}
\usepackage{color}
\usepackage[noend]{algpseudocode}
\usepackage{algorithmicx,algorithm}
\usepackage[shortlabels]{enumitem}
\usepackage{multirow}
\usepackage{colortbl}
\usepackage{subcaption}
\usepackage{rotating}
\usepackage{colortbl}
\usepackage[accsupp]{axessibility}

\input{math_commands.tex}

\usepackage[breaklinks=true,bookmarks=false]{hyperref}
\usepackage[capitalize]{cleveref}
\crefname{section}{Sec.}{Secs.}
\Crefname{section}{Section}{Sections}
\Crefname{table}{Table}{Tables}
\crefname{table}{Tab.}{Tabs.}

\iccvfinalcopy 

\def\iccvPaperID{5701} 

\ificcvfinal\fi

\begin{document}

\title{Active Neural Mapping}

\author{Zike Yan$^{1,2}$  \qquad  \qquad Haoxiang Yang$^{1,2}$ \qquad \qquad Hongbin Zha$^{1,2}$\\
	$^{1}$Key Laboratory of Machine Perception (MOE), School of EECS, Peking University\\
	$^{2}$PKU-SenseTime Machine Vision Joint Lab\\
	{\tt\small zike.yan@pku.edu.cn \qquad yyyhhhxxx@stu.pku.edu.cn \qquad zha@cis.pku.edu.cn}
}

\newcommand*{\dictchar}[1]{
	\clearpage
	\twocolumn[
	\centerline{\parbox[c][4cm][c]{10cm}{%
			\centering
			\textbf{\Large
				{#1}}}}]
}

\newcommand{\beginsupplement}{%
	\setcounter{table}{0}
	\renewcommand{\thetable}{S\arabic{table}}%
	\setcounter{figure}{0}
	\renewcommand{\thefigure}{S\arabic{figure}}%
	\setcounter{section}{0}
	\renewcommand{\thesection}{\Roman{section}}%
}

\maketitle
\ificcvfinal\thispagestyle{empty}\fi

\begin{abstract}
   We address the problem of active mapping with a continually-learned neural scene representation, namely \emph{Active Neural Mapping}. The key lies in actively finding the target space to be explored with efficient agent movement, thus minimizing the map uncertainty on-the-fly within a previously unseen environment. In this paper, we examine the weight space of the continually-learned neural field, and show empirically that the neural variability, the prediction robustness against random weight perturbation, can be directly utilized to measure the instant uncertainty of the neural map. Together with the continuous geometric information inherited in the neural map, the agent can be guided to find a traversable path to gradually gain knowledge of the environment. We present for the first time an active mapping system with a coordinate-based implicit neural representation for online scene reconstruction. Experiments in the visually-realistic Gibson and Matterport3D environment demonstrate the efficacy of the proposed method.
\end{abstract}

\input{srcs/introduction.tex}
\input{srcs/related_work.tex}
\input{srcs/preliminaries.tex}

\input{srcs/method.tex}
\input{srcs/evaluation.tex}

\section{Conclusion}
\label{sec:conclusion}
In this paper, we introduce a novel active mapping system based on implicit neural representations. The key to the solution is a goal location identification strategy through weight perturbation that drives the mobile agent to the areas with the most distribution discrepancy. The active mapping is achieved by alternatively performing action decisions to reach the goal location, and map parameter updating given incoming observations. The iterative process can be viewed as a joint optimization to reach an equilibrium point within the receding local horizon, guaranteeing a promising scene geometry recovery through autonomous exploration. The proposed strategy and the overall design of the system are justified through experiments and ablation studies. 

\subsection{Limitations and future potentials}
Though the weight perturbation provides a convenient way to find the next best view, the action decision of the system is dependent on the map-free local planner, which may occasionally get stuck by objects out of view or in narrow areas. Better exploiting the information inherited in the neural map for online navigation and replanning is one natural extension of the proposed system. Possible solutions include optimization-based planners~\cite{Adamkiewicz2022ral, Kurenkov2022ral} and INR-guided reinforcement learning to replace the existing planner trained with a 2D top-down map or raw observations. 
Meanwhile, the target view selection module simply discards other goal location candidates. Without exploiting temporal and historical cues, the agent may move back to visited areas where the complicated geometry is hard to converge. This issue may be handled by a graph model~\cite{Kai2022cvpr} for candidate organization and assignment or a decomposed and hierarchical representation for object-wise or room-wise exploration.

Enabling the mobile agent to behave autonomously in an unknown space is one straight path towards spatial intelligence~\cite{Davison2018futuremapping, Davison2019futuremapping}. The implicit neural representation has shown great potential to distill knowledge from pre-trained model~\cite{Clip-fields, Dreamfusion} for a globally consistent and informative representation. Best exploiting the information inherited in the prior and streaming data to construct a decodable and task-agnostic scene representation may lead to an innovative map-centric paradigm for the vision, graphics, and robotics communities.

\paragraph{Acknowledgements}
We thank anonymous reviewers for their fruitful comments and suggestions. This work is supported by the Joint Funds of the National Natural Science Foundation of China
(U22A2061) and National Natural Science Foundation of China (62176010).


{\small
	\bibliographystyle{ieee_fullname}
	\bibliography{egbib}
}

\newpage

\dictchar{Supplementary Materials}

\beginsupplement
\section{Runtime}
\label{sec:runtime}
Supplementary to Fig.~\ref{fig:time} of the main paper, we estimate the average portion of the computational cost induced by each module in~\cref{fig:runtime}. The majority of the runtime is distributed to the continual learning of the neural map (71\%), while best view selection and identification, visualization, and meshing take most of the rest computational resources. 

\section{Discussion on uncertainty quantification}
\label{sec:uncertain}
\begin{figure}
	\centering
	\includegraphics[width=0.98\linewidth]{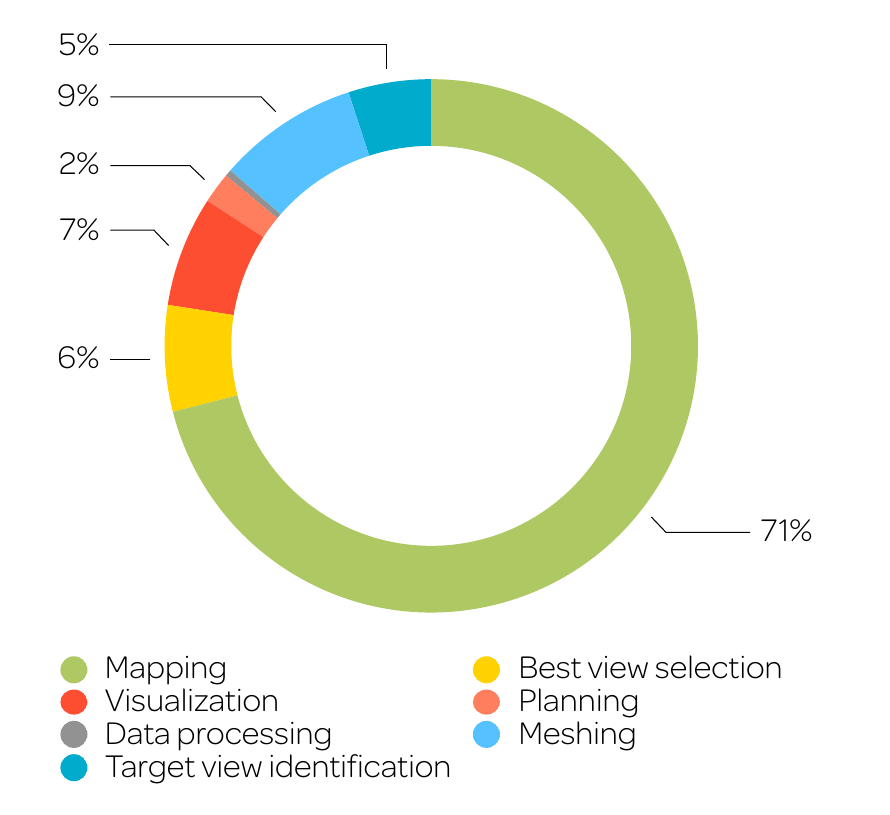}
	\caption{The runtime distribution of each module.}
	\label{fig:runtime}
\end{figure}

We refer readers to~\cite{Abdar2021review} for a more comprehensive review of uncertainty quantification methods. For active mapping, we are interested in the epistemic uncertainty that characterizes the incomplete knowledge of $\gD$ given existing training samples. In the field of implicit neural representation and learning-based active mapping, a few uncertainty quantification methods are adopted, e.g., variational inference~\cite{Shen20213dv, Shen2022eccv}, distribution analysis of the output space~\cite{Lee2022ral, Pan2022eccv}, or ensemble technique~\cite{Georgakis2022icra, Sunderhauf2022arxiv}. Variational inference based approaches reformulate the architecture and the training process, hence limiting its application to existing neural representations. Ensemble methods require random initialization for each ensemble to induce prediction variance, which is impractical for our continual learning setting. In contrast to the distribution analysis of the output space, we examine the parameter space of the implicit neural representations and formulate active mapping from a continual learning perspective. The proposed method is simple and can be easily applied to different architectures and training paradigms. The proposed problem setting is also similar to Bayesian active learning~\cite{Gal2017icml}. Hence, we also adopt the MC-Dropout layers with a BALD score as the substitute for Module 1. Compared to the proposed method, the involvement of Dropout leads to slow inference as a single forward pass leads to noisy prediction. Even though the exploration efficiency is comparable (\cref{tab:baseline_comp}), the reconstructed mesh from averaged SDF prediction is still noisy (\cref{fig:mesh-a}) even after 5 runs. Note that most modules with 98\% of the computational time in~\cref{fig:runtime} require a noise-free prediction. Increasing the number of feedforward passes would be computationally burdensome for an online system.

\section{More experimental results}
\label{sec: evaluation}
In this section, more results on the Gibson~\cite{Gibson2018cvpr} and Matterport3D datasets~\cite{Matterport2017_3DV} are provided as supplementary evaluations to Tab.~\ref{tab:baselines} and~\ref{tab:ablation} of the main paper.

\begin{figure}[t]
	\centering
	\begin{subfigure}{0.43\linewidth}
		\includegraphics[width=0.98\linewidth]{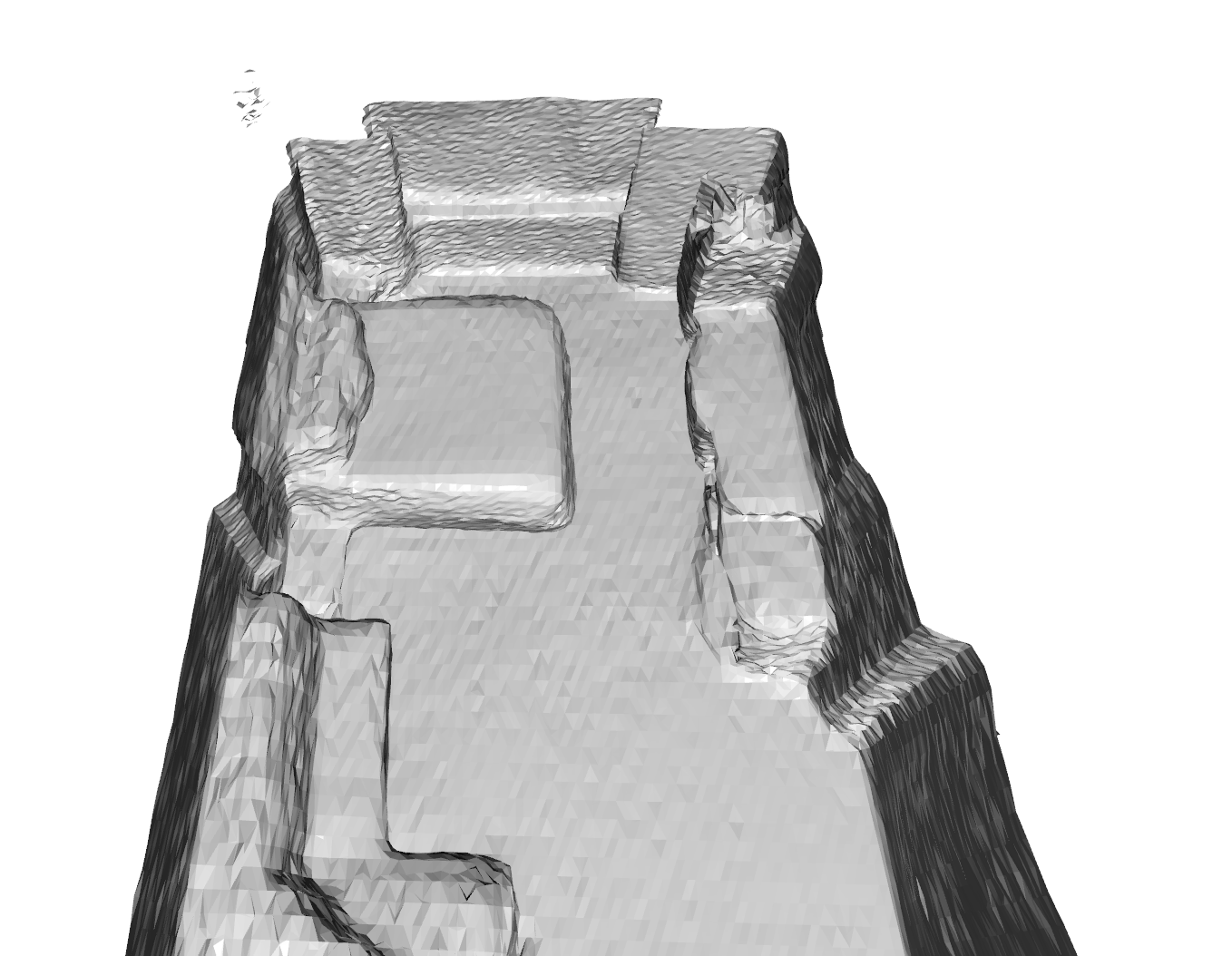}
		\caption{Module 1 (Dropout)}
		\label{fig:mesh-base-a}
	\end{subfigure}
	\begin{subfigure}{0.43\linewidth}
		\includegraphics[width=0.98\linewidth]{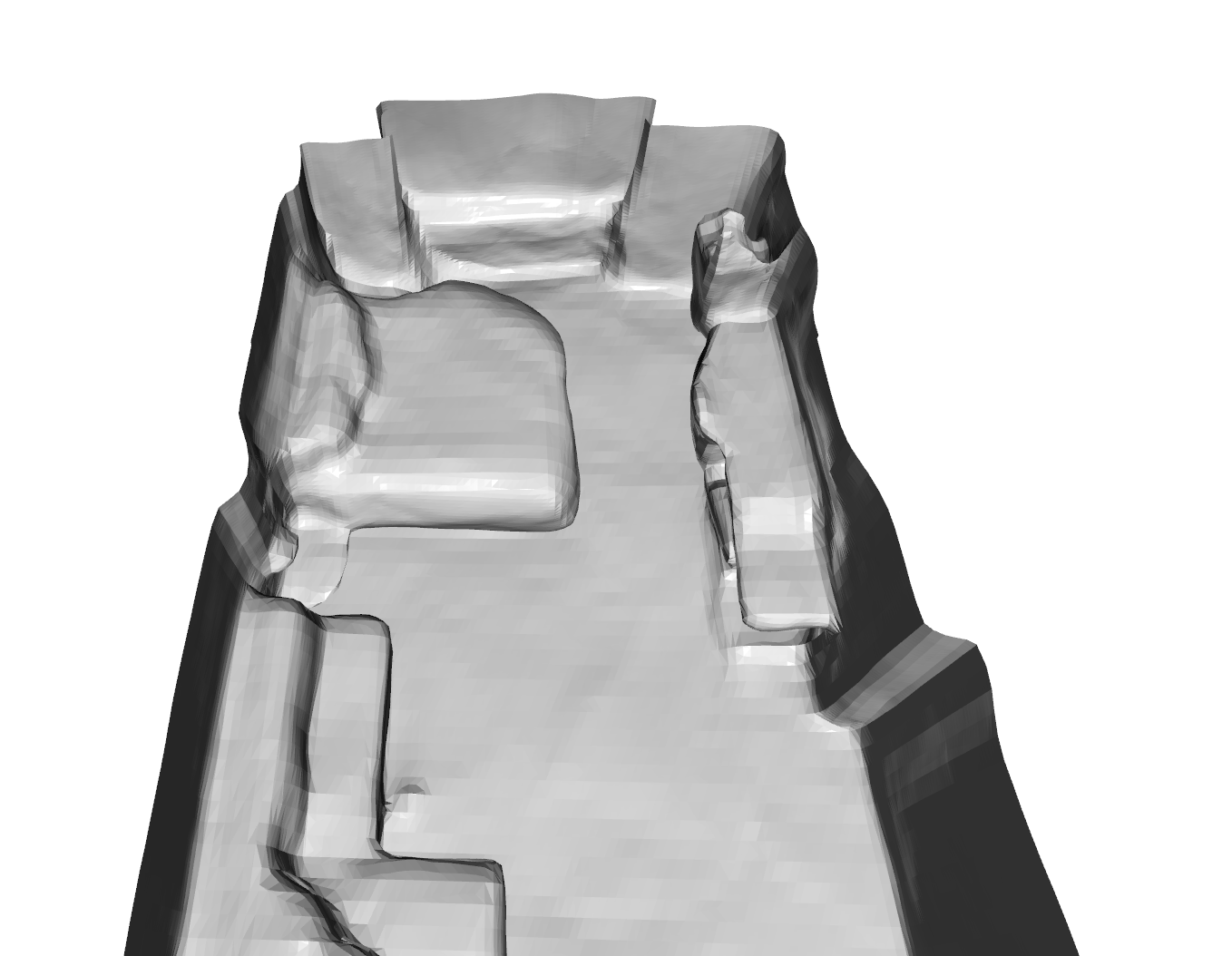}
		\caption{Module 3 (ReLU)}
		\label{fig:mesh-base-b}
	\end{subfigure}
	\begin{subfigure}{0.43\linewidth}
		\includegraphics[width=0.98\linewidth]{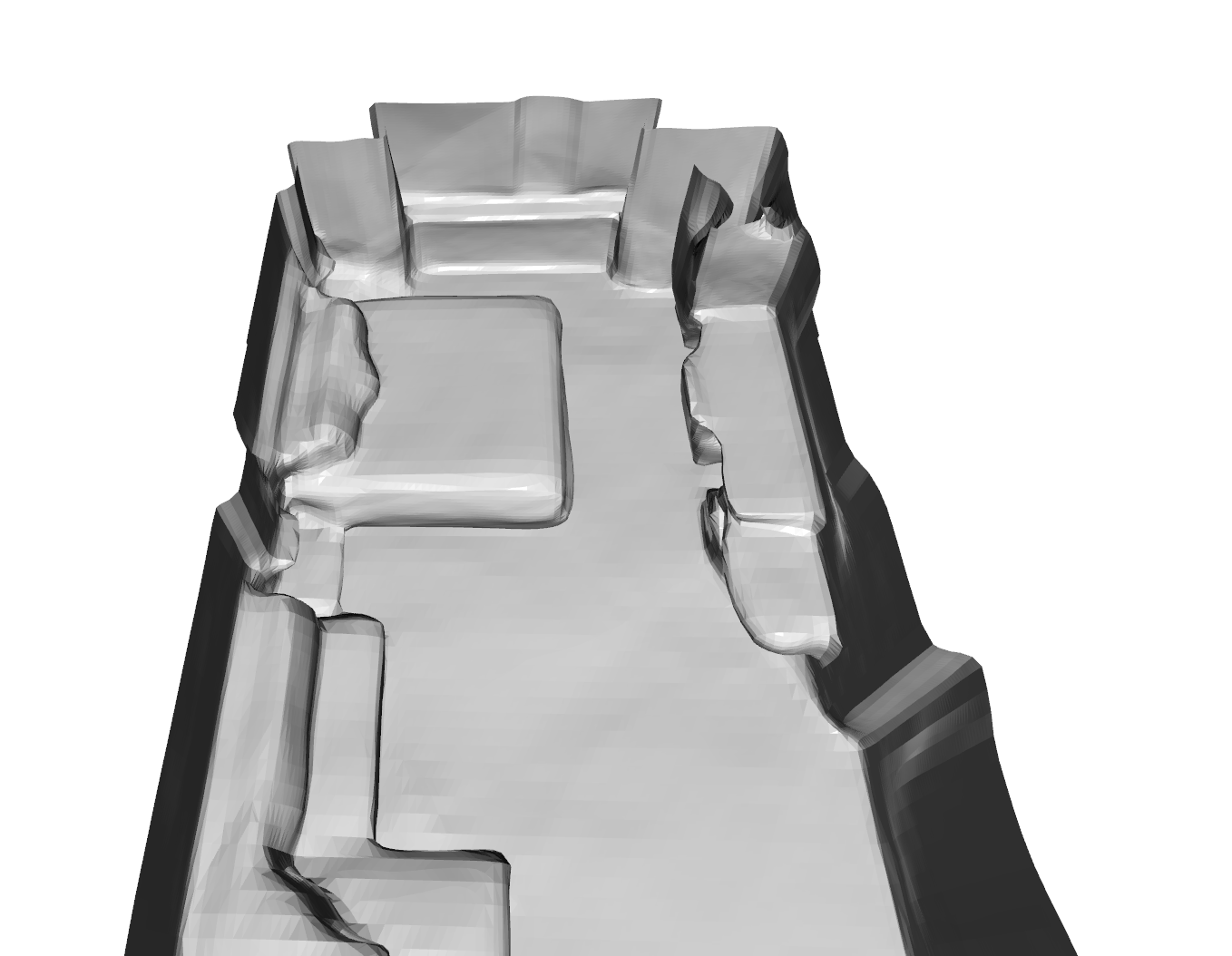}
		\caption{Final}
		\label{fig:mesh-base-c}
	\end{subfigure}
	\begin{subfigure}{0.43\linewidth}
		\includegraphics[width=0.98\linewidth]{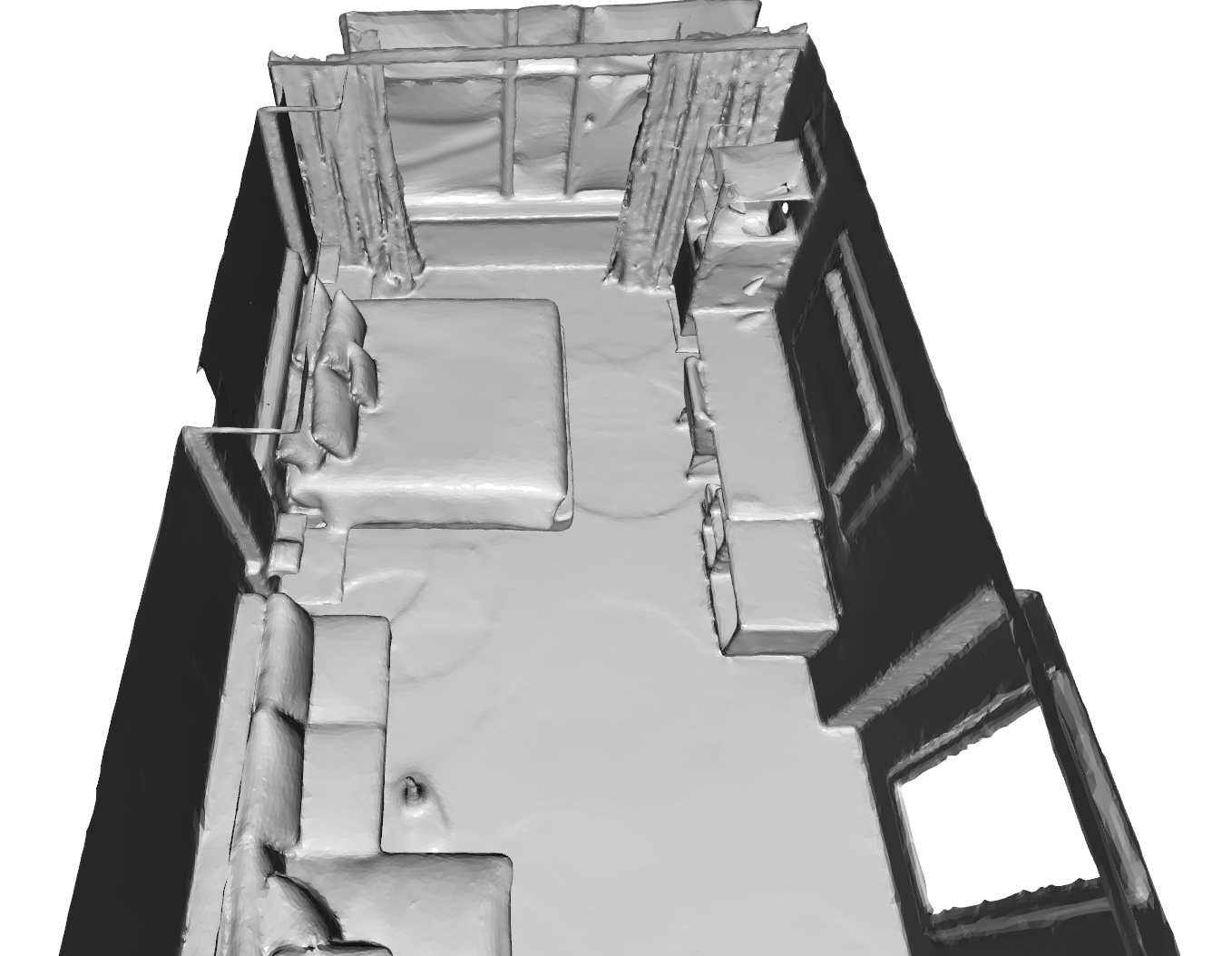}
		\caption{GT}
		\label{fig:mesh-base-d}
	\end{subfigure}
	\caption{Extracted meshes from different baselines.}
	\label{fig:baseline_meshes}
\end{figure}

\subsection{Per-scene evaluation}
\label{subsec:per_scene}
We conduct experiments on single-floor scenes with less than 10 rooms. Scenes in Matterport3D are consistently larger and more complex than in Gibson (see the number of keyframes (Kfs) in~\cref{tab:method_comp}). The keyframes are automatically selected as frames with high prediction losses. Both a large scene that requires a long trajectory for exploration and a complex scene that can not be easily recovered lead to a large number of keyframes, hence indicating the scene complexity. For small scenes (num\_room $<=$ 5), the active mapping is conducted in 1000 steps; While for large scenes (num\_room $>$ 5), we run 2000 steps for better coverage. $<$ 5cm (\%) is defined as the ratio of ground truth mesh vertices whose predicted distance value is within 5cm. Acc. is defined as the mesh-wise mean distance between the estimated one and the ground truth.

\begin{figure}[t]
	\centering
	\begin{subfigure}{0.32\linewidth}
		\includegraphics[width=0.98\linewidth]{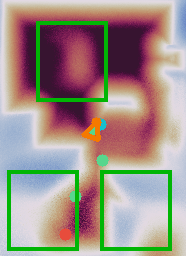}
		\caption{Module 1}
		\label{fig:2d-base-a}
	\end{subfigure}
	\begin{subfigure}{0.32\linewidth}
		\includegraphics[width=0.98\linewidth]{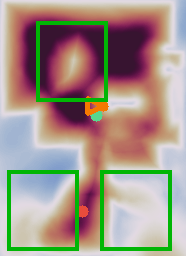}
		\caption{Module 3}
		\label{fig:2d-base-b}
	\end{subfigure}
	\begin{subfigure}{0.32\linewidth}
		\includegraphics[width=0.98\linewidth]{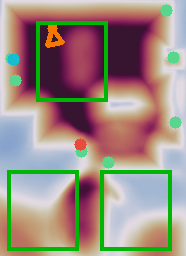}
		\caption{Final}
		\label{fig:2d-base-c}
	\end{subfigure}
	\caption{Learned SDF from different baselines.}
	\label{fig:baseline_2d}
\end{figure}

From the per-scene evaluation, we can obtain a better understanding of the proposed method. When comparing to the relevant methods in~\cref{tab:method_comp}, we can find a typical failure case of HxpkQ, which will be further explained in~\cref{subsec:limitation}. The major reason for the failure is the local planner that gets stuck in the narrow area. Meanwhile, the evaluations on different module substitutes in~\cref{tab:baseline_sdf}-\ref{tab:baseline_comp} uncover several important insights: 

\noindent$\bullet$ Trade-offs between mapping accuracy and area coverage. As a surface-based exploration method, the choice of different network architectures and penalty functions may lead to diverse performance. For instance, the ReLU activations without skip-connection lead to inferior accuracy in high-frequency areas (see~\cref{fig:mesh-base-b}), hence the target view identification module may guide the agent to visited areas as the high-frequency geometric details are not recovered well. Consequently, the mesh quality and the exploration completeness decline (See~\cref{tab:baseline_mesh} and \ref{tab:baseline_comp}). 

\begin{table*}
	\footnotesize
	\centering
	\caption{Comparison against relevant methods regarding the completeness of actively-captured observations.}
	\label{tab:method_comp}
	\begin{tabular}{@{}lc!{}lc!{}lc!{}lc!{}lc!{}lc!{}} 
		\toprule
		&&&\multicolumn{3}{c!{}}{ \textbf{Comp. (\%) ↑}}&\\
		& \textbf{Rooms} & \textbf{Kfs} & \textbf{Random} & \textbf{FBE} & \textbf{UPEN} & \textbf{OccAnt} & \textbf{Ours} \\
		\toprule
		\textbf{Gibson}-Cantwell* & 8 & 38 & 24.43 & 40.93 & 39.42 & 37.96 & \textbf{61.36} \\ 
		\hline
		\textbf{Gibson}-Denmark & 2 & 18 & 27.83 & 70.28 & 66.41 & 65.07 & \textbf{85.86} \\ 
		\hline
		\textbf{Gibson}-Eastville* & 6 & 34 & 14.32 & 58.49 & 51.51 & 27.03 & \textbf{74.21} \\ 
		\hline
		\textbf{Gibson}-Elmira & 3 & 14 & 66.29 & 72.69 & 82.14 & 84.37 & \textbf{91.65} \\ 
		\hline
		\textbf{Gibson}-Eudora & 3 & 13 & 53.89 & 76.65 & 75.74 & 74.07 & \textbf{90.12} \\ 
		\hline
		\textbf{Gibson}-Greigsville & 2 & 23 & 75.44 & 90.34 & 73.72 & 88.27 & \textbf{92.47} \\ 
		\hline
		\textbf{Gibson}-Pablo & 4 & 14 & 46.87 & \textbf{76.06} & 54.16 & 64.50 & 72.88\\ 
		\hline
		\textbf{Gibson}-Ribera & 3 & 14 & 44.29 & 79.26 & 81.21 & 66.97 & \textbf{88.62} \\ 
		\hline
		\textbf{Gibson}-Swormville* & 7 & 31 & 58.81 & 55.46 & 45.43 & 48.64 & \textbf{66.86} \\ 
		\hline
		\textbf{Gibson}-mean & 4 & 22 & 45.80  & 68.30 & 63.30 &61.88 & \textbf{80.45}\\ 
		\midrule
		\textbf{MP3D}-GdvgF* & 6 & 32 & 68.45 & 81.78 & \textbf{82.39} & 80.24 & 80.99 \\ 
		\hline
		\textbf{MP3D}-gZ6f7 & 1 & 48 & 29.81 & 81.01 & 82.96  & \textbf{82.02} & 80.68 \\ 
		\hline
		\textbf{MP3D}-HxpKQ* & 8 & 32 & 46.93 & 58.71 & 52.70 & \textbf{60.50} & 48.34 \\ 
		\hline
		\textbf{MP3D}-pLe4w & 2 & 52 & 32.92 & 66.09 & 66.76 &  67.13 & \textbf{76.41} \\ 
		\hline
		\textbf{MP3D}-YmJkq & 4 & 68 & 50.26 & 68.32 & 60.47 & 68.70 & \textbf{79.35} \\ 
		\hline
		\textbf{MP3D}-mean &4 & 46 & 45.67 & 68.53 & 69.09 & 71.72 &\textbf{73.15}\\ 
		\bottomrule
	\end{tabular}
	\hspace{0.2em}
	\begin{tabular}{ccccc} 
		\toprule
		&\multicolumn{3}{c!{}}{ \textbf{Comp. (cm)↓}}&\\
		\textbf{Random} & \textbf{FBE} ↑ & \textbf{UPEN} & \textbf{OccAnt} & \textbf{Ours} \\
		\toprule
		59.59& 37.03 & 42.12 & 43.27 &\textbf{17.67} \\ 
		\hline
		50.42& 12.40 & 17.34 & 16.96 & \textbf{3.78} \\ 
		\hline
		72.39 & 24.08 & 28.16 & 60.67 & \textbf{11.36} \\ 
		\hline
		11.63 & 10.40 & 5.35 & 4.35 & \textbf{2.57} \\ 
		\hline
		23.24 & 8.11 & 9.18 & 9.27 & \textbf{2.27} \\ 
		\hline
		6.97 & 2.62 & 16.34 & 3.25 & \textbf{1.78} \\ 
		\hline
		34.70 & \textbf{6.38} & 31.81 & 20.49 & 9.96 \\ 
		\hline
		33.27 & 6.53 & 5.74 & 18.67 & \textbf{4.13} \\ 
		\hline
		18.10 & 22.19 & 33.78 & 32.34 & \textbf{13.43} \\ 
		\hline
		34.48 & 14.42 & 21.09 & 23.25 & \textbf{7.44}\\
		\midrule
		11.67 & 5.48 & \textbf{5.14}& 5.66 & 5.69 \\ 
		\hline
		46.48 & 7.06 & \textbf{6.14} & 6.19 & 7.43\\ 
		\hline
		19.10 & \textbf{11.75} & 14.11& \textbf{11.75} & 15.96 \\ 
		\hline
		30.79 & 12.78 & 11.82& 11.51 & \textbf{8.03} \\ 
		\hline
		24.61 & 11.85 & 15.77 & 11.90 & \textbf{8.46} \\ 
		\hline
		26.53 & 9.78 & 10.60 & 9.40 &\textbf{9.11}\\
		\bottomrule
	\end{tabular}
\end{table*}

\begin{figure*}
	\centering
	\begin{subfigure}{0.18\linewidth}
		\includegraphics[width=0.98\linewidth]{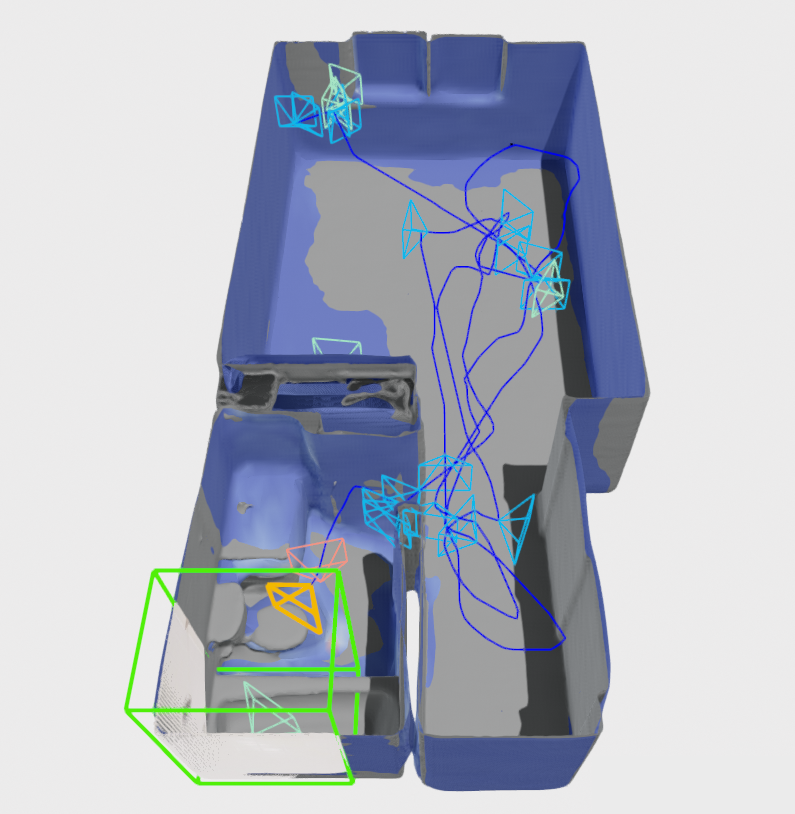}
		\caption{Edgemere}
		\label{fig:mesh-a}
	\end{subfigure}
	\begin{subfigure}{0.175\linewidth}
		\includegraphics[width=0.98\linewidth]{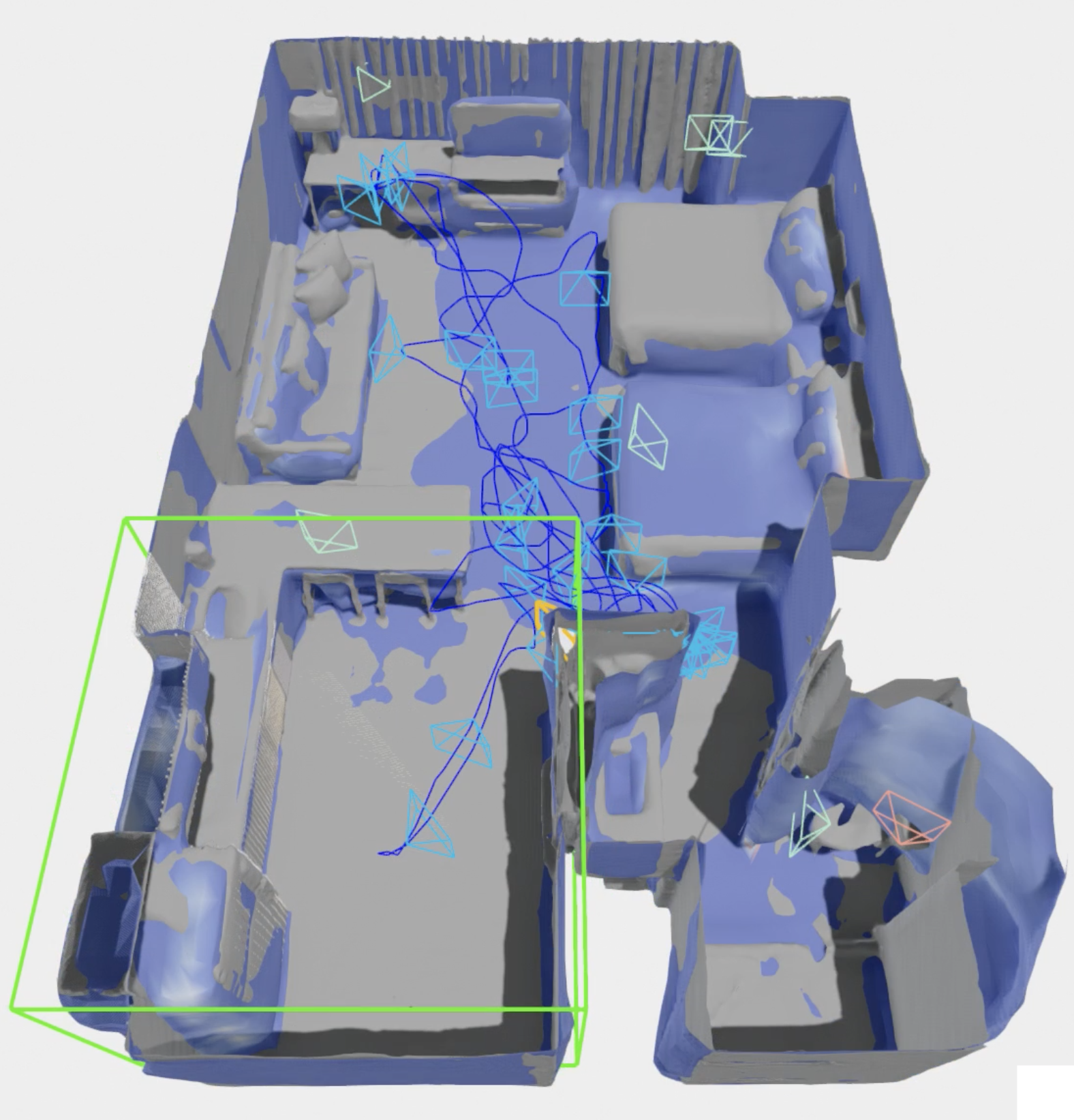}
		\caption{Elmira}
		\label{fig:mesh-b}
	\end{subfigure}
	\begin{subfigure}{0.265\linewidth}
		\includegraphics[width=0.98\linewidth]{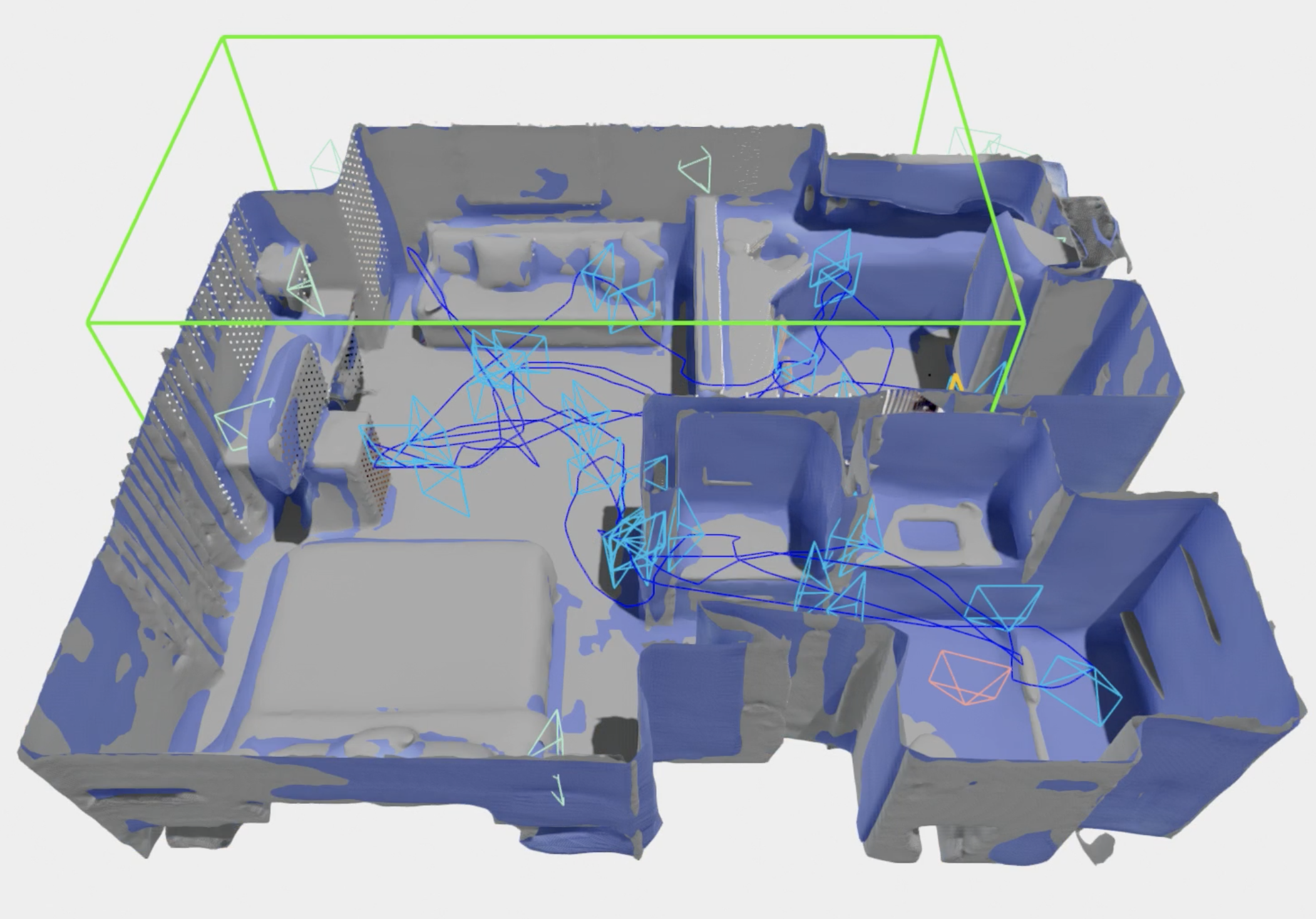}
		\caption{Eudora}
		\label{fig:mesh-c}
	\end{subfigure}
	\begin{subfigure}{0.305\linewidth}
		\includegraphics[width=0.98\linewidth]{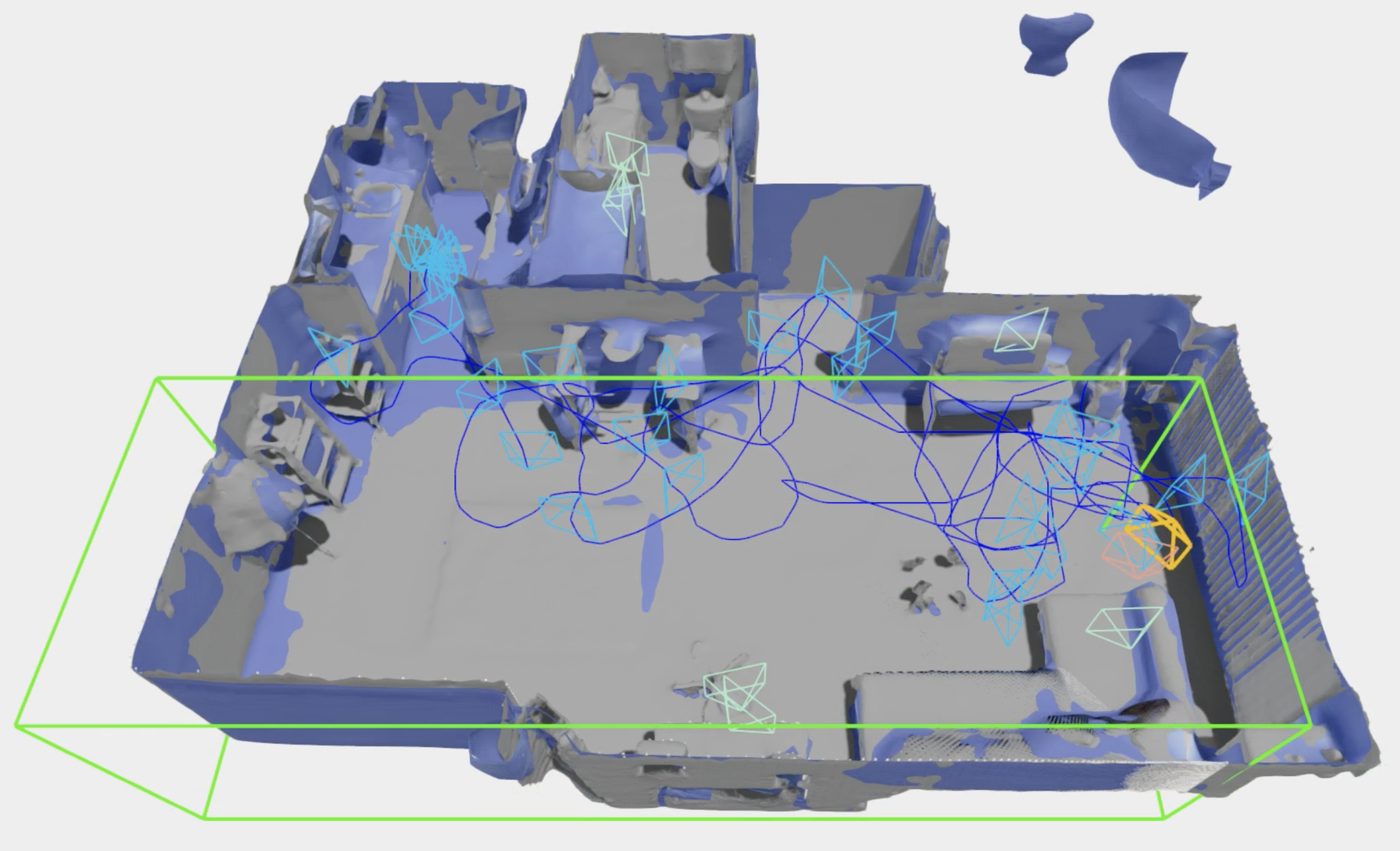}
		\caption{Greigsville}
		\label{fig:mesh-d}
	\end{subfigure}
	\caption{Examples of well-explored scenes. The ground truth mesh is in grey while the estimated one is in blue.}
	\label{fig:good_examples}
\end{figure*}

\begin{figure}
	\centering
	\begin{subfigure}{0.4\linewidth}
		\includegraphics[width=0.98\linewidth]{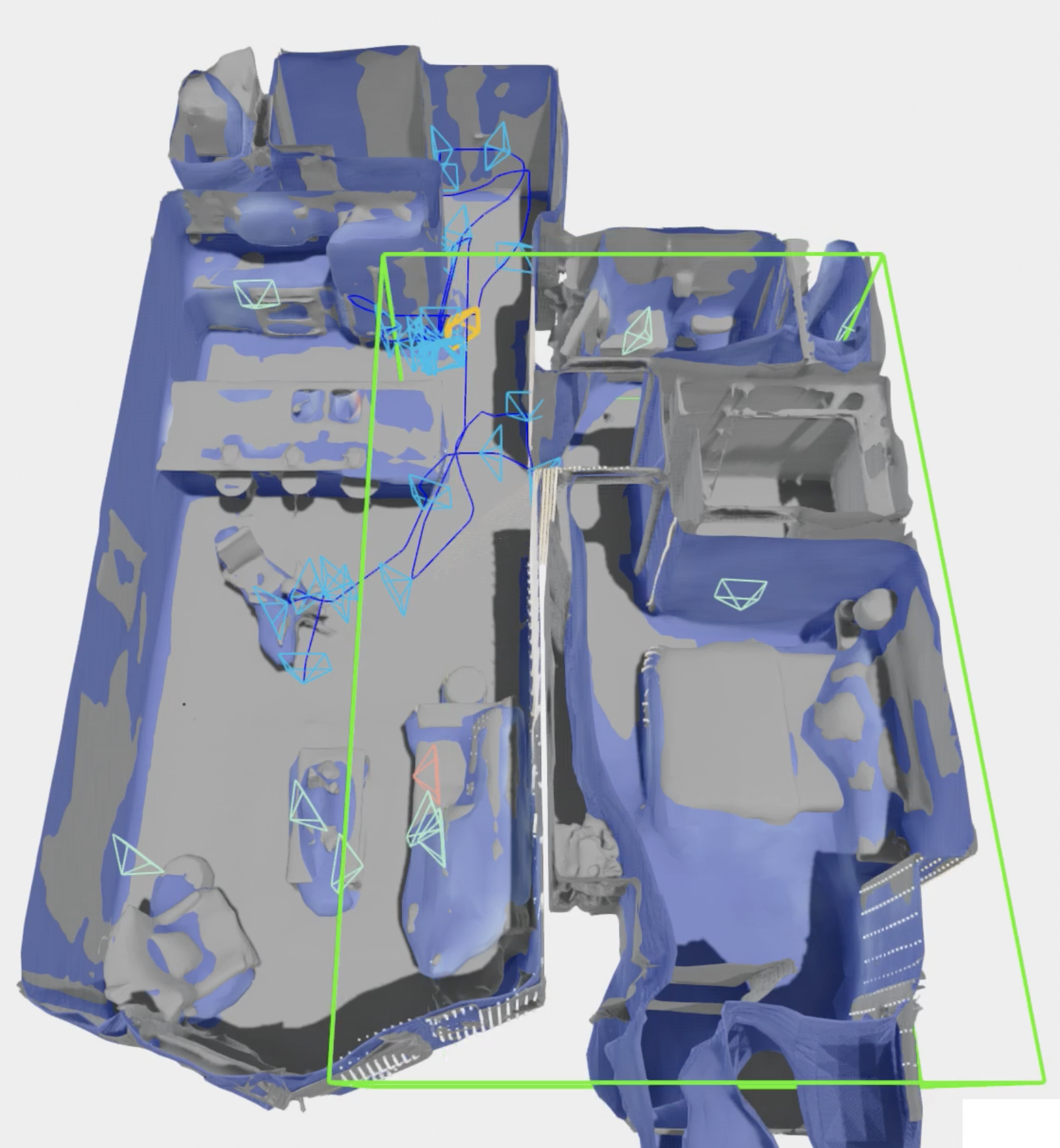}
		\caption{Gibson-Swormville}
		\label{fig:fail-a}
	\end{subfigure}
	\begin{subfigure}{0.41\linewidth}
		\includegraphics[width=0.98\linewidth]{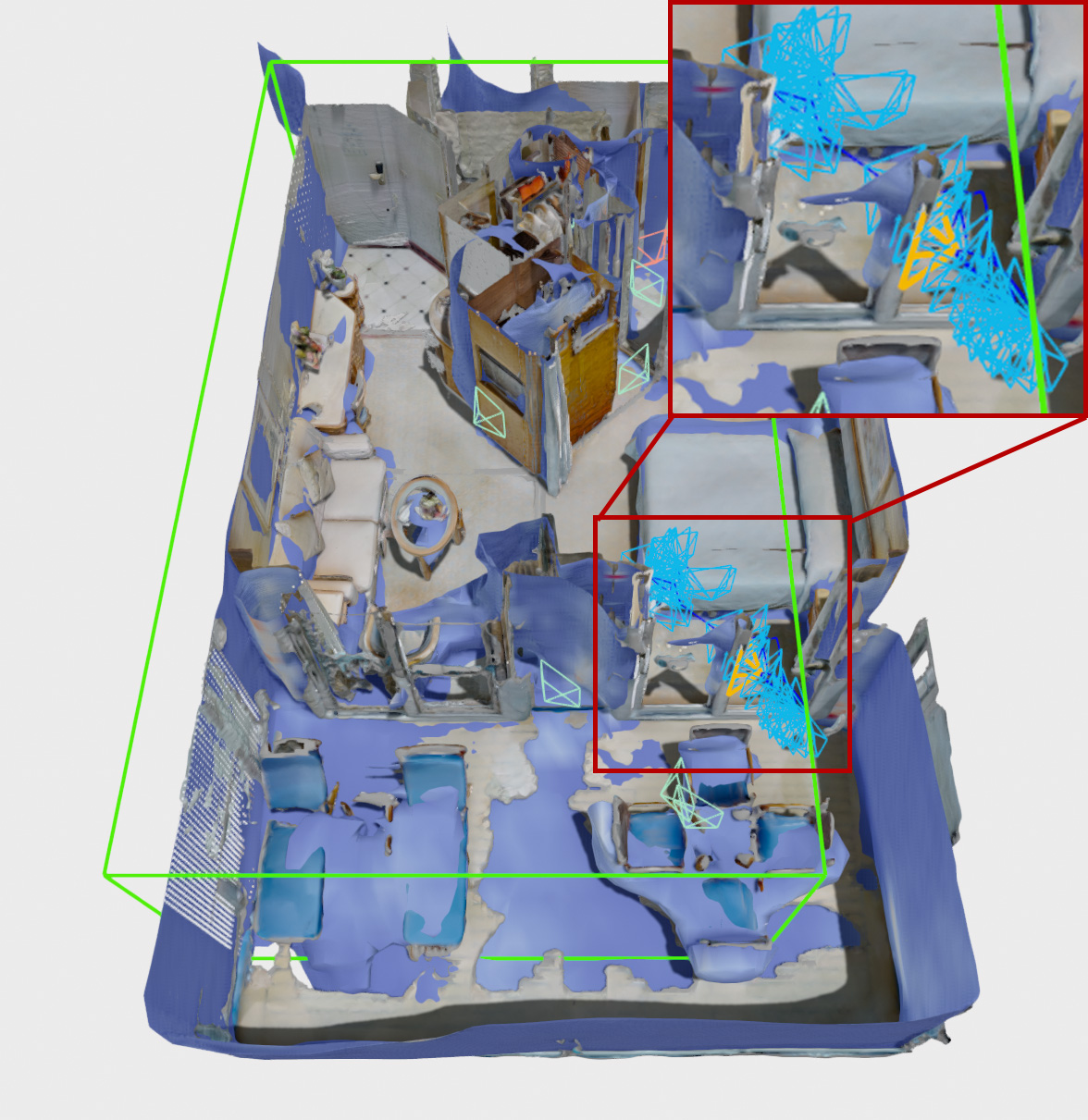}
		\caption{MP3D-HxpKQ}
		\label{fig:fail-d}
	\end{subfigure}
	\begin{subfigure}{0.87\linewidth}
		\includegraphics[width=0.98\linewidth]{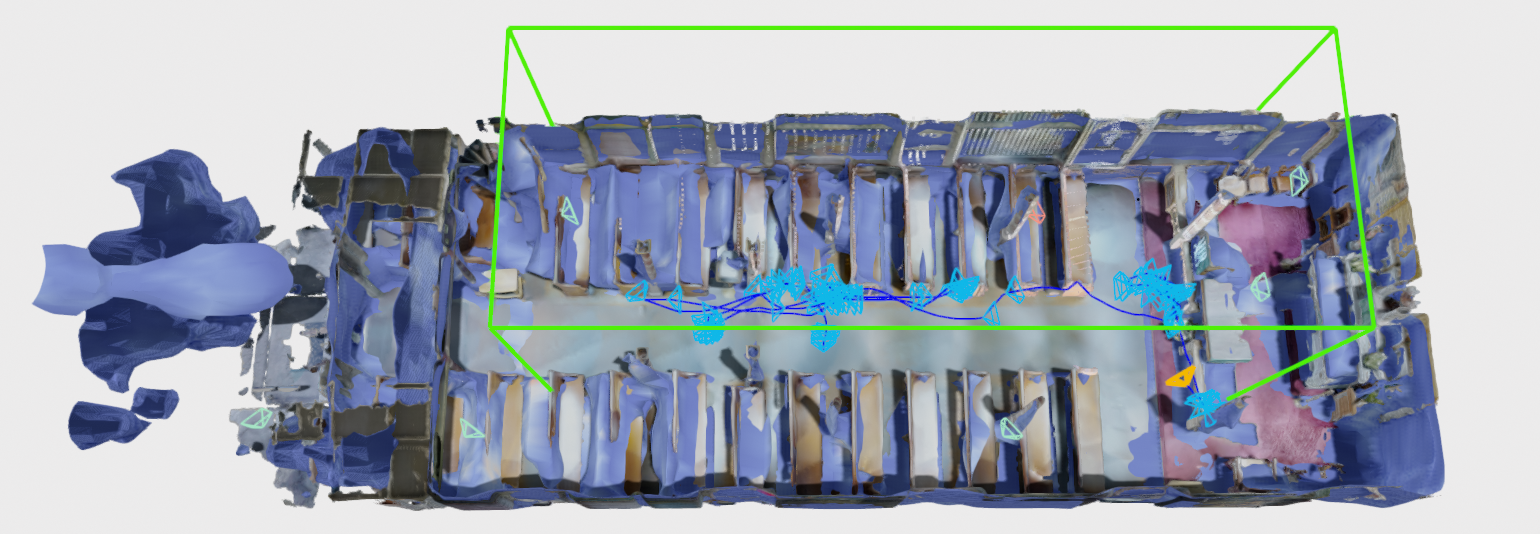}
		\caption{MP3D-YmJkq}
		\label{fig:fail-c}
	\end{subfigure}
	\caption{Typical challenges: (a) low $<$5cm(\%) due to incomplete exploration; (b) low Comp. as the agent gets stuck; (c) high FPR due to extreme occlusions.}
	\label{fig:failure_cases}
\end{figure}

\noindent$\bullet$ Generalization ability matters. Meanwhile, there are also different prediction behaviors in unobserved areas when deployed under different architectures/penalties, leading to different generalization abilities. As illustrated in~\cref{fig:baseline_2d}, the predicted distance values in unexplored areas (within the green bounding box) are smaller for module 3 baseline (PosEnc+ReLU) than the module 1 (PosEnc+SoftPlus+Dropout) and final baselines (PosEnc+SoftPlus). This may affect the generation of false-positive zero-crossing surfaces and the convergence rate when new data come, thus affecting the active mapping performance. That's also the reason why module 4 achieves higher $<$5cm (\%) in~\cref{tab:baseline_sdf} even if it explores fewer areas (\cref{tab:baseline_comp}) with a worse mesh model (\cref{tab:baseline_mesh}).

\subsection{Limitations and typical challenges}
\label{subsec:limitation}
As illustrated in~\cref{fig:good_examples}, the proposed method can recover high-fidelity scene geometry through autonomous exploration. All rooms are well-explored, thus achieving high Comp. in~\cref{tab:method_comp}. Compared to other baselines, the proposed method achieves much better coverage. It should be noted that the Comp. metrics are only affected by the action (trajectory). The results clearly verify the efficacy of the proposed method.

We here emphasize the typical challenges that lead to unsatisfactory results. As illustrated in~\cref{fig:fail-a}, the bad SDF prediction results mainly contribute to incomplete exploration. On the other hand, the mesh quality is highly relevant to the complexity of the scene geometry. As illustrated in~\cref{fig:fail-c}, although a straight path leads to efficient exploration, the church scenario is full of occlusions that make accurate reconstruction non-trivial. Finally, as illustrated in~\cref{fig:fail-d}, as the agent gets stuck all the time, all metrics are unpleasant for HxpKQ even if the environment is open without occlusions. 

We can also find that the major reason for the failure cases is the unsatisfactory trajectory/navigation. Even though target location candidates exist in the unexplored rooms (see~\cref{fig:fail-a}), entering a narrow door/corridor or even going upstairs brings little information gain sometimes due to a limited field of view. If the agent gets stuck easily in nearby areas, the target candidate may never be chosen as the optimal one. For future extensions, better local planning strategies should be deployed to handle the challenges as discussed in Sec. 6.1 of the main paper.

\clearpage
\begin{table*}
	\scriptsize
	\centering
	\caption{Comparison against baselines regarding the SDF prediction accuracy of the zero-crossing surface.}
	\label{tab:baseline_sdf}
	\begin{tabular}{@{}lc!{}lc!{}lc!{}lc!{}lc!{}lc!{}} 
		\toprule
		&&&\multicolumn{3}{c!{}}{ \textbf{MAD (cm) ↓}}&\\
		& \textbf{Rooms} & \textbf{Kfs} & \textbf{Random} & \textbf{Module 1} & \textbf{Module 3} & \textbf{Module 4} & \textbf{Final} \\
		\toprule
		\textbf{Gibson}-Cantwell* & 8 & 38 & 8.83 & 6.10 & 8.12 & 8.28 & \textbf{5.20} \\ 
		\hline
		\textbf{Gibson}-Denmark & 2 & 18 & 9.06 & 4.59 & 4.74 & 4.92 & \textbf{4.32} \\ 
		\hline
		\textbf{Gibson}-Eastville* & 6 & 34 & 16.66 & 6.65 & \textbf{5.19} & 6.96 & 6.64 \\ 
		\hline
		\textbf{Gibson}-Elmira & 3 & 14 & 4.72 & 5.01 & \textbf{3.89} & \textbf{3.89} & 4.01 \\ 
		\hline
		\textbf{Gibson}-Eudora & 3 & 13 & 8.03 & 4.22 & 4.62 & 7.08 & \textbf{4.02} \\ 
		\hline
		\textbf{Gibson}-Greigsville & 2 & 23 & 4.55 & 4.10 & 4.35 & \textbf{3.87} & 4.18 \\ 
		\hline
		\textbf{Gibson}-Pablo & 4 & 14 & 7.48 & \textbf{4.87} & 6.66 & 7.96  & 5.23\\ 
		\hline
		\textbf{Gibson}-Ribera & 3 & 14 & 6.29 & 5.27 & \textbf{4.95} & 5.79 & 5.13 \\ 
		\hline
		\textbf{Gibson}-Swormville* & 7 & 31 & 10.80 & 10.29 & 6.46 & \textbf{6.22} & 7.17 \\ 
		\hline
		\textbf{Gibson}-mean & 4 & 22 & 8.49 & 5.68 & 5.42 & 6.10 & \textbf{5.10}\\ 
		\midrule
		\textbf{MP3D}-GdvgF* & 6 & 32 & 6.78 & 6.67 & 7.53 & 5.97 & \textbf{3.77} \\ 
		\hline
		\textbf{MP3D}-gZ6f7 & 1 & 48 & 23.12 & 4.12 & 5.37 &  4.46 & \textbf{3.18} \\ 
		\hline
		\textbf{MP3D}-HxpKQ* & 8 & 32 & \textbf{4.60} & 5.69 & 7.31 & 4.99 & 7.03 \\ 
		\hline
		\textbf{MP3D}-pLe4w & 2 & 52 & 6.22 & 3.14 & 4.48 &  \textbf{2.83} & 3.25 \\ 
		\hline
		\textbf{MP3D}-YmJkq & 4 & 68 & 3.65 & 3.61 & 5.60 &  \textbf{3.25} & 4.22 \\ 
		\hline
		\textbf{MP3D}-mean & 4 & 46 &8.87&4.65&6.06&4.30&\textbf{4.29}\\ 
		\bottomrule
	\end{tabular}
	\begin{tabular}{ccccc} 
		\toprule
		&\multicolumn{3}{c!{}}{ \textbf{$<$5cm (\%) ↑}}&\\
		\textbf{Random} & \textbf{Module 1}  & \textbf{Module 3} & \textbf{Module 4} & \textbf{Final} \\
		\toprule
		39.07 & 51.23 & 46.52 & 45.95 &\textbf{59.93} \\ 
		\hline
		45.68 & 68.47 & 66.98 & 65.22 &\textbf{70.14} \\ 
		\hline
		15.50 & 46.72 & \textbf{56.95} & 42.95 & 51.33 \\ 
		\hline
		64.65 & 63.04 & 70.84 & 71.42 &\textbf{72.27} \\ 
		\hline
		51.20 & 66.48 & 66.88 & 55.68 &\textbf{71.98} \\ 
		\hline
		66.52 & 71.61 & 68.38 & 71.10 &\textbf{74.95} \\ 
		\hline
		50.49 & \textbf{62.16} & 45.36 & 55.35 & 60.41 \\ 
		\hline
		59.59 & 58.42 & 63.51 & 58.66 & \textbf{65.37} \\ 
		\hline
		28.40 & 32.60 & 49.16 & \textbf{53.29} & 44.16 \\ 
		\hline
		46.79&57.86&59.40&57.74&\textbf{63.39}\\
		\midrule
		67.27 & 70.23 & 63.39 & 76.74 & \textbf{77.05} \\ 
		\hline
		28.01 & \textbf{77.59} & 64.42 & 75.42 & 75.03 \\ 
		\hline
		67.80 & 61.47 & 49.52 & \textbf{66.97} & 53.72 \\ 
		\hline
		63.60 & 78.79 & 64.47 & \textbf{82.78} & 75.22 \\ 
		\hline
		72.44 & 74.59 & 55.56 & \textbf{82.12} & 72.67 \\ 
		\hline
		59.82 &72.53&59.47&\textbf{76.81}&70.74\\
		\bottomrule
	\end{tabular}
\end{table*}
\begin{table*}
	\scriptsize
	\centering
	\caption{Comparison against baselines regarding the reconstructed mesh quality.}
	\label{tab:baseline_mesh}
	\begin{tabular}{@{}lc!{}lc!{}lc!{}lc!{}lc!{}lc!{}} 
		\toprule
		&&&\multicolumn{3}{c!{}}{ \textbf{FPR (\%) ↓}}&\\
		& \textbf{Rooms} & \textbf{Kfs} & \textbf{Random} & \textbf{Module 1} & \textbf{Module 3} & \textbf{Module 4} & \textbf{Final} \\
		\toprule
		\textbf{Gibson}-Cantwell* & 8 & 38 & 51.32 & 60.80 & \textbf{37.37} & 51.95 & 45.41 \\ 
		\hline
		\textbf{Gibson}-Denmark & 2 & 18 & 40.43 & 17.13 & \textbf{15.83} & 22.96 & 22.31 \\ 
		\hline
		\textbf{Gibson}-Eastville* & 6 & 34 & 68.04 & 54.89 & \textbf{44.86} & 61.70 & 46.33 \\ 
		\hline
		\textbf{Gibson}-Elmira & 3 & 14 & 25.02 & 33.54 & 16.71 & 24.54 & \textbf{15.94} \\ 
		\hline
		\textbf{Gibson}-Eudora & 3 & 13 & 22.47 & 17.69 & \textbf{17.56} & 23.57 & 18.06 \\ 
		\hline
		\textbf{Gibson}-Greigsville & 2 & 23 & 26.63 & 27.80 & 24.18  & 26.12 & \textbf{23.07} \\ 
		\hline
		\textbf{Gibson}-Pablo & 4 & 14 & 28.82 & 24.43 & 21.99 & \textbf{20.31} & 22.51 \\ 
		\hline
		\textbf{Gibson}-Ribera & 3 & 14 & 27.25 & \textbf{20.53} & 23.08 & 28.88 & 34.51 \\ 
		\hline
		\textbf{Gibson}-Swormville* & 7 & 31 & 39.17 & 55.04 & \textbf{24.21} & 35.72 & 24.26 \\ 
		\hline
		\textbf{Gibson}-mean & 4 & 22 &36.57&34.65&\textbf{25.09}&32.86&28.04\\ 
		\midrule
		\textbf{MP3D}-GdvgF* & 6 & 32 & 25.55 & 27.32 & 22.74 & 32.09 & \textbf{22.68} \\ 
		\hline
		\textbf{MP3D}-gZ6f7 & 1 & 48 & 69.45 & \textbf{33.12} & 36.68 & 40.37 & 33.53 \\ 
		\hline
		\textbf{MP3D}-HxpKQ* & 8 & 32 & 48.67 & 46.31 & 47.78 & 60.29 & \textbf{45.83} \\ 
		\hline
		\textbf{MP3D}-pLe4w & 2 & 52 & 50.02 & 36.54 & \textbf{31.69} & 35.00 & 35.76 \\ 
		\hline
		\textbf{MP3D}-YmJkq & 4 & 68 & 65.70 & 71.86 & \textbf{56.34} & 72.20 & 62.54 \\ 
		\hline
		\textbf{MP3D}-mean & 4 & 46 &51.88&43.03&\textbf{39.05}&47.99&40.07\\ 
		\bottomrule
	\end{tabular}
	\hspace{0.1em}
	\begin{tabular}{ccccc} 
		\toprule
		&\multicolumn{3}{c!{}}{ \textbf{Acc. (cm)↓}}&\\
		\textbf{Random} & \textbf{Module 1} & \textbf{Module 3} & \textbf{Module 4} & \textbf{Final} \\
		\toprule
		13.83 & 17.24 & \textbf{7.97} & 13.67 & 9.70 \\ 
		\hline
		8.37 & \textbf{3.63} & 3.91 & 4.56 & 4.76 \\ 
		\hline
		34.60 & 29.33 & 24.54 & 34.96 & \textbf{20.39} \\ 
		\hline
		5.19 & 6.56 & 3.77 & 5.27 & \textbf{3.74} \\ 
		\hline
		4.67 & \textbf{3.58} & 3.84 & 5.26 & 3.99 \\ 
		\hline
		12.44 & 14.09 & 8.50 & \textbf{6.35} & 9.63 \\ 
		\hline
		8.39 & 5.86 & 4.78 & \textbf{4.71} & 6.34 \\ 
		\hline
		6.02 & \textbf{5.52} & 5.77 & 11.26 & 11.52 \\ 
		\hline
		10.69 & 29.16 & 6.11 & 15.73 &\textbf{5.10} \\ 
		\hline
		11.58&12.77&\textbf{7.69}&11.31&8.35\\
		\midrule
		6.55 &  6.45& 5.93 & 8.12 &\textbf{5.09} \\ 
		\hline
		19.57 & 7.89 & 10.00 & 9.13 &\textbf{4.15} \\ 
		\hline
		12.43 & 16.13 & \textbf{12.28} & 16.70 & 15.60 \\ 
		\hline
		11.15 & 8.01 &  6.66 & 7.39 & \textbf{5.56} \\ 
		\hline
		37.58& 42.12 & 28.28 & 53.24 & \textbf{8.61} \\ 
		\hline
		17.46&16.12&12.63&18.92&\textbf{7.80}\\
		\bottomrule
	\end{tabular}
\end{table*}
\begin{table*}
	\scriptsize
	\centering
	\caption{Comparison against baselines regarding the completeness of actively-captured observations.}
	\label{tab:baseline_comp}
	\begin{tabular}{@{}lc!{}lc!{}lc!{}lc!{}lc!{}lc!{}} 
		\toprule
		&&&\multicolumn{3}{c!{}}{ \textbf{Comp. (cm) ↓}}&\\
		& \textbf{Rooms} & \textbf{Kfs} & \textbf{Random} & \textbf{Module 1} & \textbf{Module 3} & \textbf{Module 4} & \textbf{Final} \\
		\toprule
		\textbf{Gibson}-Cantwell* & 8 & 38 & 59.59 & \textbf{13.96} & 66.23 & 41.84 & 17.67 \\ 
		\hline
		\textbf{Gibson}-Denmark & 2 & 18 & 50.42 & \textbf{1.86} & 2.70 & 3.00 & 3.78 \\ 
		\hline
		\textbf{Gibson}-Eastville* & 6 & 34 & 72.39 & 14.32 & 21.44 & 11.28 & \textbf{11.36} \\ 
		\hline
		\textbf{Gibson}-Elmira & 3 & 14 & 11.63 & 2.87 & \textbf{1.38} & 3.97 & 2.57 \\ 
		\hline
		\textbf{Gibson}-Eudora & 3 & 13 & 23.24 & \textbf{2.05} & 2.62 & 2.43 & 2.27 \\ 
		\hline
		\textbf{Gibson}-Greigsville & 2 & 23 & 6.97 & 1.87 & \textbf{0.86} & 1.37 & 1.78 \\ 
		\hline
		\textbf{Gibson}-Pablo & 4 & 14 & 34.70 & \textbf{5.47} & 13.92 & 22.16 & 9.96 \\ 
		\hline
		\textbf{Gibson}-Ribera & 3 & 14 & 33.27 & 4.97 & 4.32 & 15.99 & \textbf{4.13} \\ 
		\hline
		\textbf{Gibson}-Swormville* & 7 & 31 & 18.10 & 20.92 & \textbf{12.85} & 17.50 & 13.43 \\ 
		\hline
		\textbf{Gibson}-mean & 4 & 22 &34.48&7.59&14.04&13.28&\textbf{7.44}\\ 
		\midrule
		\textbf{MP3D}-GdvgF* & 6 & 32 & 11.67 & \textbf{5.13} & 4.92 & 7.67 & 5.69 \\ 
		\hline
		\textbf{MP3D}-gZ6f7 & 1 & 48 & 46.48 & 7.53 & 8.94 & 8.63 & \textbf{7.43} \\ 
		\hline
		\textbf{MP3D}-HxpKQ* & 8 & 32 & 19.10 & \textbf{13.85} & 14.69 & 30.37 & 15.96 \\ 
		\hline
		\textbf{MP3D}-pLe4w & 2 & 52 & 30.79 & 7.51 & \textbf{6.14} & 5.15 & 8.03 \\ 
		\hline
		\textbf{MP3D}-YmJkq & 4 & 68 & 24.61 & 15.05 & \textbf{8.41} & 12.01 & 8.46 \\ 
		\hline
		\textbf{MP3D}-mean & 4 & 46 &26.53&9.78&\textbf{8.62}&12.77&9.11\\
		\bottomrule
	\end{tabular}
	\hspace{0.1em}
	\begin{tabular}{ccccc} 
		\toprule
		&\multicolumn{3}{c!{}}{ \textbf{Comp. (\%) ↑}}&\\
		\textbf{Random} & \textbf{Module 1} & \textbf{Module 3} & \textbf{Module 4} & \textbf{Final} \\
		\toprule
		24.43 & \textbf{68.77} & 23.07 & 39.19 &61.36 \\ 
		\hline
		27.83 & \textbf{93.46} & 88.90 & 88.00 & 85.86 \\ 
		\hline
		14.32 & 57.67 & 66.21 & 73.67 &\textbf{74.21} \\ 
		\hline
		66.29 & 89.03 & \textbf{95.04} & 86.04 & 91.65 \\ 
		\hline
		53.89 & \textbf{91.57} & 90.29 & 90.06 & 90.12 \\ 
		\hline
		75.44 & 91.32 & \textbf{97.79} & 94.70 & 92.47 \\ 
		\hline
		46.87 & \textbf{79.74} & 70.50 & 63.97 & 72.88 \\ 
		\hline
		44.29 & 85.86 & 87.01 & 67.98 &\textbf{88.62} \\ 
		\hline
		58.81 & 57.92 & \textbf{66.88}& 59.73 & 66.86 \\ 
		\hline
		45.80 &79.48&76.19&73.70&\textbf{80.45}\\
		\midrule
		68.45 & 82.14 & \textbf{82.51} & 76.75 &80.99 \\ 
		\hline
		29.81 & \textbf{80.99} & 79.06 & 76.75 & 80.68 \\ 
		\hline
		46.93 & 49.71 & \textbf{51.26} & 34.32 & 48.34 \\ 
		\hline
		32.92 & 78.91 & \textbf{83.02} & 86.10 & 76.41 \\ 
		\hline
		50.26 & 65.32 & 77.29 & 64.85 &\textbf{79.35} \\ 
		\hline
		45.67 & 71.41 & \textbf{74.63} &67.75&73.15\\
		\bottomrule
	\end{tabular}
\end{table*}

\end{document}

%% file: math_commands.tex
\usepackage{amsmath,amsfonts,bm}

\def\eqref#1{equation~\ref{#1}}

\def\1{\bm{1}}

\def\vtheta{{\bm{\theta}}}
\def\va{{\bm{a}}}

\def\vu{{\bm{u}}}
\def\vv{{\bm{v}}}

\def\vx{{\bm{x}}}
\def\vy{{\bm{y}}}
\def\vz{{\bm{z}}}

\DeclareMathAlphabet{\mathsfit}{\encodingdefault}{\sfdefault}{m}{sl}
\SetMathAlphabet{\mathsfit}{bold}{\encodingdefault}{\sfdefault}{bx}{n}

\def\gD{{\mathcal{D}}}

\def\gL{{\mathcal{L}}}

\def\gX{{\mathcal{X}}}
\def\gY{{\mathcal{Y}}}

\def\sR{{\mathbb{R}}}

\newcommand{\E}{\mathbb{E}}
\newcommand{\Ls}{\mathcal{L}}



%% file: srcs/introduction.tex
\section{Introduction}
\label{sec:intro}
 How we represent a 3D environment accurately and efficiently is of tremendous importance for vision, robotics, and graphics communities. Recent advances in implicit neural representations (INRs) cast the issue as a low-dimensional function regression problem. Parameterized by a single network $\vtheta$, the quantity of interest $\vy$ such as color, occupancy, and semantics can be efficiently queried with the spatial coordinates $\vx$ through a feedforward pass $\vy = f(\vx;\vtheta)$. Unlike traditional representations that discretize the entire space and explicitly store a set of the input-output samples $\{\vx_i, \vy_i\}_N$ in manually-designed data structures such as voxel grid, surfel, and triangle mesh, the implicit neural representation is proved to have great capacity~\cite{Tancik2020nips, Fathony2020iclr, Yuce2022cvpr} that recovers complex signals at a constant small size, guaranteeing high-fidelity view synthesis~\cite{Mildenhall2020eccv, Barron2021iccv, Mildenhall2022cvpr} and accurate geometry reconstruction~\cite{Sitzmann2020nips, Lindell2022cvpr, Azinovic2022cvpr}. Nonetheless, the quality of the learned implicit neural representation is highly data-dependent: as implicit neural representations are trained through self-supervision given discrete training samples, insufficient sampling frequency leads to geometric and texturing artifacts~\cite{Shen2021_3dv, Yu2021cvpr, Yuce2022cvpr, Shen2022eccv}. 

 \begin{figure}[t]
 	\centering
 	\includegraphics[width=0.98\linewidth]{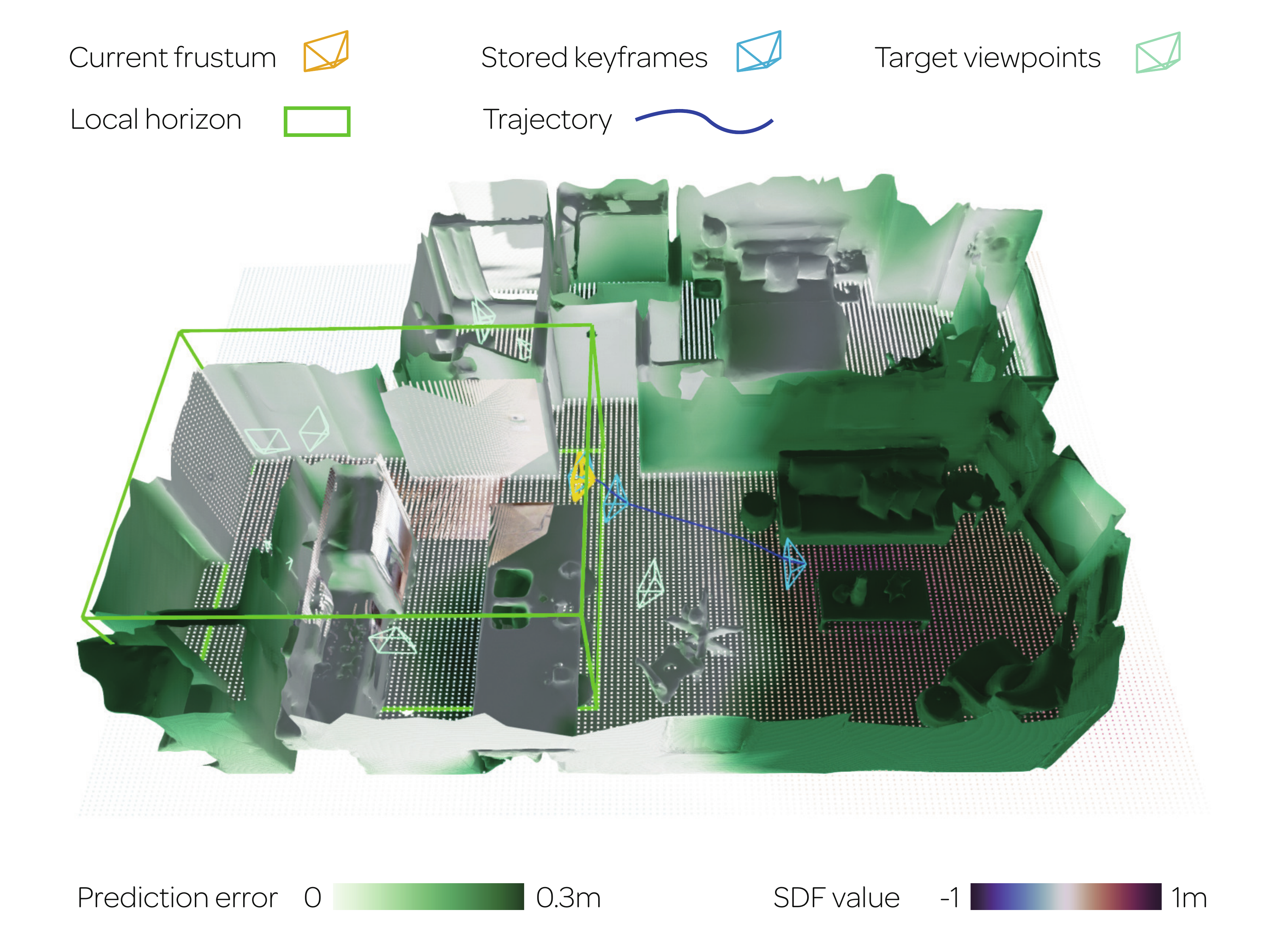}	
 	\caption{An overview of the proposed active neural mapping system. Guided by the continually-updated neural map (visualized as the SDF values through a forward pass), the mobile agent explores the environment actively to minimize the prediction uncertainty (visualized as the prediction error given the truth surface points). }
 	\label{fig:onecol}
 \end{figure}

Unlike conventional methods that rely on passive data acquisition, we address the problem of \emph{active neural mapping}, where a 3D neural field is constructed on the fly with an actively-exploring mobile agent to best represent the scene. The target is to find efficient agent movement within the previously-unknown environment to gradually minimize the map uncertainty. Similar problems such as autonomous exploration and next-best-view planning are well-studied~\cite{Connolly1985icra, Maver1993pami, Yamauchi1997frontier, Whaite1997pami, Placed2022survey} by exploiting discretized scene representations to achieve the best coverage and reconstruction accuracy (see Sec.~\ref{sec:related_work} for a detailed discussion). Though the implicit neural representation has its own merits, \eg promising representation power and continuous/differentiable properties, this problem setting poses new challenges to the INRs: new knowledge of the environment is actively captured and constantly distilled to the neural map, where the neural map is expected to 1) specify the uncertain areas to be explored; 2) provide reliable geometric information for reconstruction; 3) allow for incremental updating given constantly observed data.
 
 In this paper, we show for the first time a continual learning perspective of online active mapping based on the coordinate-based implicit neural representation. Inspired by the seminal works of~\cite{Sucar2021iccv, Yan2021iccv}, we adopt the incremental updating of a continuous neural signed distance field. The key to our active mapping solution lies in a novel uncertainty quantification manner of the learned neural map through weight perturbation. We show empirically that the replayed buffer during continual learning forces the neural network to land in a low-loss basin given previously observed data to avoid forgetting, while resulting in a sharp ridge given erroneously-generated zero-crossings from not-well-explored areas to ensure transferability. That is to say, as the weight changes constantly during continual learning, the robustness of the predicted signed distance values exhibit distinguishable behaviors against weight perturbations for explored and unexplored surface samples. These findings share similar spirits with recent studies in neuroscience~\cite{Ma2006NN, Orban2016neuron, Festa2021neuronal} and learning theory~\cite{Xie2021anv, Shi2021nips}, allowing us to explicitly reason the uncertain areas within the neural field and guide the mobile agent for actively capturing new information.
 
 Our active mapping system adopts ideas from both frontier-based and sampling-based exploration strategies. The neural variabilities of zero-crossing samples are examined under random weight perturbations, where samples with high variation are viewed as target areas to be explored. Along with the continually-learned geometric information, the neural map guides the agent to explore the environment actively. The key contributions can be summarized as follows:

$\bullet$ We provide a new perspective of active mapping from the optimization dynamics of map parameters.

$\bullet$ We introduce an effective online active mapping system in a continual learning fashion.

$\bullet$ We propose a novel uncertainty quantification manner through weight perturbation for goal location identification.

%% file: srcs/related_work.tex
\section{Related Work}
\label{sec:related_work}
\paragraph{Active mapping.}
Active mapping aims to find the optimal sensor movements to capture observations that best represent a scene, thus minimizing the uncertainty of the environment through exploration. Typical approaches can be categorized into frontier-based and sampling-based ones from a goal location identification perspective. Frontier-based methods explore by approaching the selected frontier (regions on the boundary between the explored free space and the unexplored space~\cite{Yamauchi1997frontier}), aiming to push the boundary of the explored areas until the entire space is observed. The major differences lie in the frontier detection strategies~\cite{Zhu2015ictai, Dornhege2013ar, Shen2012icra, Zhou2021ral, Batinovic2021ral} and the best frontier selection strategies~\cite{Dai2020icra, Cieslewski2017iros, Holz2010isr, Kai2022cvpr}. 

On the other hand, sampling-based methods adopt random or guided sampling of potential viewpoints in the workspace and incrementally grow a Rapidly-exploring Random Tree (RRT)~\cite{Lavalle1998rrt} or a Rapidly-exploring Random Graph (RRG)~\cite{Karaman2010rss} to find the traversable paths. The next best view is repeatedly selected along the best branch in a \emph{receding horizon} fashion~\cite{Bircher2016icra} to maximize a given objective function. Unlike frontier-based methods that focus more on the map coverage, sampling-based methods allow different objectiveness, \eg, localization uncertainty~\cite{Papachristos2017icra}, geometric uncertainty~\cite{Schmid2020ral, Ramakrishnan2020eccv, Georgakis2022icra}, visual saliency~\cite{Dang2018icra}, and vehicle dynamics~\cite{Dharmadhikari2020icra}, to be taken into account.

To take advantage of both frontier-based and sampling-based methods, new strategies are employed in a hybrid or an informed sampling-based fashion. The hybrid method~\cite{Selin2019ral} adopts a sampling-based manner for local planning, while utilizing a frontier-based method for global planning to handle the dead-end case as sampling-based methods can easily get stuck locally. Meanwhile, as most computational resources are wasted on the redundant utility computation of non-selected samples~\cite{Song2018icra, Batinovic2022ral}, the informed sampling based methods~\cite{Kompis2021ral, Meng2017ral, Respall2021icra} are proposed that sample candidates around frontiers to ensure faster exploration.
\vspace{-2em}
\paragraph{Dense metric representations.}
Dense metric representations play important roles in path planning as they provide complete geometric information for any queried location within the workspace. Existing active mapping methods mainly rely on the volumetric representation that discretizes the space into voxel grids. Occupancy grid map, for example, allows distinguishing between free, occupied, and unknown space. Most occupancy grid based methods are deployed in 2D~\cite{Yamauchi1997frontier,Chaplot2019iclr,Gupta2017cvpr} for tractable computation as a mobile device typically moves at a constant height~\cite{Kaufman2018irc}. There are also 3D extensions~\cite{Bircher2016icra,Dornhege2013ar, Batinovic2021ral} that exploit an Octomap structure~\cite{Hornung2013octomap} for recursive updating of the occupancy status. Meanwhile, it is noted that merely occupancy information may be insufficient for certain gradient-based planners, CHOMP~\cite{Zucker2013ijrr} for instance. Therefore, the Euclidean signed distance field (ESDF) is introduced to be updated incrementally from a truncated signed distance field (TSDF)~\cite{Oleynikova2017iros,Pan2022iros} or a 3D occupancy grid map~\cite{Han2019iros} using Breadth-First Search (BFS), allowing online planning on a CPU-only platform.

Recent advances in implicit neural representations (INRs)~\cite{Mildenhall2020eccv, Park2019cvpr,Mescheder2019cvpr,Chen2019cvpr} facilitate multiple robotics-related downstream tasks. By encoding the coordinate-based scene properties in weights of a neural network, INRs are able to recover fine-grained scene properties with light-weight parameters~\cite{Sitzmann2020nips, Tancik2020nips, Fathony2020iclr,Yuce2022cvpr}. Hence, accurate scene geometry can be recovered with a single network~\cite{Azinovic2022cvpr, Lindell2022cvpr}. On the other hand, the gradient can be efficiently extracted from the continuous neural field through automatic differentiation. Together with the geometric information, a smooth trajectory can be optimized for collision avoidance~\cite{Adamkiewicz2022ral, Kurenkov2022ral}. Recently,~\cite{Lee2022ral,Pan2022eccv}share a similar idea of refining a coarsely-trained NeRF by actively selecting new viewpoints for batch retraining. Inspired by the continual learning fashion of online neural field updating~\cite{Yan2021iccv,Sucar2021iccv,Zhu2022cvpr,Ortiz2022rss}, we extend the works to an online active mapping framework, where the implicit neural field is updated on the fly to guide the exploration for complete coverage and constant uncertainty reduction. There are also two concurrent works~\cite{Ran2023ral,Zeng2023icra} that are most related to ours, tackling the inward view selection and path planning for object reconstruction. 

%% file: srcs/preliminaries.tex
\section{Preliminaries}
\label{sec:pre}
Given an indoor environment as the workspace $\gX \in \sR^3$ that is unknown \emph{a prior}, we aim to best represent the scene property of interest\footnote{In this paper, we target a continuous signed distance function to represent the scene surfaces.} $\gY \in \sR^m$ with a continuous function parameterized by a single MLP $\vtheta$, establishing the mapping $f(\vx;\vtheta): \gX \to \gY$ between spatial coordinates $\vx\in\gX$ and the corresponding scene property $\vy \in \gY$. To obtain an optimal map representation, a mobile agent is deployed to actively capture sensory data $\{\vx_i, \vy_i\}^t \sim \vz^t \subset \gD$ sampled from the scene surfaces $\gD$ (depth sequence in our case) with self-decided control $\va^t$ at each time, and the map parameters $\vtheta$ is updated incrementally with incoming observations.

From a global optimum view, the map can be optimized through empirical risk minimization given a pre-defined penalty function $\Ls$ and sufficient samples from the true distribution of $\gD$ as:
\begin{equation}
	\vtheta^*=\arg\min \E_{(\vx,\vy)\sim\gD}(\Ls(\vx,\vy; \vtheta)).
	\label{eq:global}
\end{equation}

In our case of an online setting, the continual learning of the map can be cast as minimizing a cumulative loss~\cite{Raghavan2021nips} within a time interval $[t,t+k]$ in the following steps as:
\begin{equation}
	\vtheta^t = \arg\min\sum_{\tau=t}^{t+k}\lambda^\tau\E_{(\vx^t,\vy^t)\sim\vz^{1:\tau}}(\Ls(\vx^t,\vy^t; \vtheta^t)),
	\label{eq:cumulative}
\end{equation}
where the observation $\vz^t$ is conditioned on past controls $\va^{1:t}$, and $k\rightarrow\infty$ equals an unending exploration setting.

From~\cref{eq:cumulative}, we can see that the overarching goal of an optimal map is intractable to be achieved as future observations $\vz^{t:t+k}$ are not available. This issue is formalized by Raghavan and Balaprakash~\cite{Raghavan2021nips} from a generalization-forgetting perspective. They point out that the penalty $\gL$ is needed to not only prevent catastrophic forgetting of previous observations, but improve generalization to new data. As proved in~\cite{Raghavan2021nips}, the dynamics of continual learning are affected by three factors: the cost of prediction error over all past observations $\E_{(\vx,\vy)\sim\vz^{1:t}}(\Ls(\vx,\vy; \vtheta^t))$, the cost variation arising from the data distribution shift $\Delta \vx^t$, and the cost variation arising from the parameter changes $\Delta \vtheta^t$:
\begin{equation}
	\begin{aligned}
		H(\delta \vz,\vtheta^t) &\approx \beta L(\vtheta^t,\vz^{1:t}) \\ 
		&+ \sum_{\tau=t}^{t+k}(L(\vtheta^\tau,\vz^{1:\tau}\cap\delta\vz) - L(\vtheta^\tau,\vz^{1:\tau})) \\ 
		&+ \sum_{\tau=t}^{t+k}(L(\vtheta^\tau+\delta\vtheta,\vz^{1:\tau}) - L(\vtheta^\tau,\vz^{1:\tau})), 
		\label{eq:tradeoff}
	\end{aligned}
\end{equation}
where $L(\vtheta^i,\vz^{j}) = \E_{(\vx^j,\vy^j)\sim\vz^{j}}(\Ls(\vx^j,\vy^j; \vtheta^i))$. 

Intuitively, minimal $H(\delta \vz,\vtheta^t)$ induced by distribution shift $\delta\vz$ and parameter changes $\delta \vtheta$ indicates that the arrival of a new observation does not affect the current optimal solution $\vtheta^t$, thus achieving the global optimum. Even though such an optimum cannot be guaranteed, a saddle point\footnote{Given that $H(\delta\vz,\vtheta^*) \leq H(\delta\vz^*,\vtheta^*) \leq H(\delta\vz^*,\vtheta)$, the equilibrium point of $\{\delta\vz^*, \vtheta^*\}$ can be found by alternatively updating the data discrepancy to maximize the generalization, and then optimizing $\vtheta$ to avoid forgetting given the maximum generalization.} can be found. In~\cite{Raghavan2021nips}, the discrepancy between two subsequent tasks is maximized, followed by the minimization of forgetting under the maximum generalization. This manner lays the theoretic foundation for us to solve the active mapping problem: if we iteratively find the most distribution shift of $\delta \vz$ and update the map parameters $\vtheta$ given a new observation, we converge to a local equilibrium point within the small time interval $k$ according to \cref{eq:tradeoff}.

The optimization perspective of \cref{eq:cumulative,eq:tradeoff} well distinguishes the proposed problem setting, namely \emph{active neural mapping}, from previous research. Recent INR-based passive SLAM systems~\cite{Sucar2021iccv, Zhu2022cvpr, Ortiz2022rss} or multi-view stereopsis methods~\cite{Mildenhall2020eccv, Azinovic2022cvpr, Zhi2021iccv} merely minimize the first term in~\cref{eq:tradeoff}, while we further take the agent action optimization into account to serve as a local generalization maximizer. Consequently, the actively captured training samples can better mimic the actual distribution $\gD$ compared to the passive observations $\vz^{1:t}$. Compared to traditional active mapping methods, we explicitly conduct map optimization through back-propagation instead of the heuristically-designed fusion techniques. The goal location is decided in a data-driven manner (see \cref{sec:method} for details) instead of the ad-hoc goal location identification strategies. The objectiveness of active mapping in~\cref{eq:tradeoff} allows for continual and lifelong ($t\to\infty$) optimization even when the agent stops, while INR-based planners~\cite{Adamkiewicz2022ral, Kurenkov2022ral, Lee2022ral} target navigating to the specified location in a pre-built or batch-optimized map. Finally, compared to recent works of object reconstruction~\cite{Lee2022ral,Pan2022eccv} that autonomously refine a pre-built coarse map $\vtheta^0$ through inward-facing view selection and planning, we target a more challenging case to incrementally optimize the map $\vtheta$ in a scene-level from scratch.
%

%% file: srcs/method.tex
\section{Active Neural Mapping}
\label{sec:method}
\begin{figure}[t]
	\begin{subfigure}{0.48\linewidth}
		\includegraphics[width=0.98\linewidth]{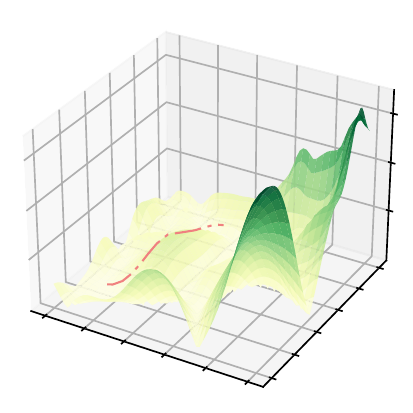}
		\caption{$|f(\vx^+;\vtheta)|$}
	\end{subfigure}
	\begin{subfigure}{0.48\linewidth}
		\includegraphics[width=0.98\linewidth]{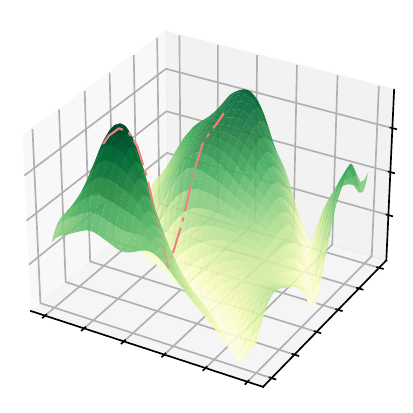}
		\caption{$|f(\vx^-;\vtheta)|$}
	\end{subfigure}
	\caption{The loss landscape $|f(\vx;\vtheta(u,v))|$ evaluated on a true surface point $\vx^+$ and a false-positive point $\vx^-$. The pink dotted lines indicate the actual loss variation along the continually learned $\vtheta^{1:T}$. It is clear that the landscape of the true surface point stays in a low-loss basin, while that of the false-positive point falls along a sharp ridge that reaches the low-loss valley once.}
	\label{fig:loss_landscape}
\end{figure}

As noticed in~\cref{sec:pre}, central to our method is the identification of the next target view that brings a significant distribution shift $\delta\vz^*$. A local planner is then deployed as a generalization maximizer that decides the following agent movement $\va^{t:t+k}$ to the target location and captures the corresponding data. Given the locally upper-bounded $H(\delta\vz^*,\vtheta^t)$, the map parameters are optimized with the new observation, thus achieving a local equilibrium point of $H(\delta\vz^*,\vtheta^*)$. The process is iteratively conducted that drives the mobile agent to actively explore the environment. In this section, we begin with an empirical analysis of how $\delta\vz^*$ can be found. The implementation of the active neural mapping system is introduced afterward.

\subsection{Through the lens of loss landscape}
\cref{eq:tradeoff} motivates us to understand the behavior of the loss $L(\vtheta, \vz)$ during continual learning: the equilibrium point of $\{\delta\vz^*, \vtheta^*\}$ indicates the requirement of a flat low-loss landscape for surface points to avoid forgetting (the minimization of the first and third terms in~\cref{eq:tradeoff}) and an evident loss discrepancy for finding $\delta\vz$ so the generalization is maximized. Following~\cite{Li2018nips, Verwimp2021iccv}, we define a hyperplane by two orthonormal vectors $\{\vu, \vv\}$,\footnote{We choose the initial and the final weights during continual learning as $\vtheta^1$ and $\vtheta^3$, and train another network with the same initialization as $\vtheta^2$. The orthonormal vectors can be obtained by orthogonalizing and normalizing the two basis vectors $(\vtheta^2-\vtheta^1, \vtheta^3-\vtheta^1)$} where any sample $\vtheta$ in the weight space can be represented by the linear combination of the two vectors as $\vtheta(u,v) = u\vu + v\vv$. We can then estimate the prediction $f(\vx; \vtheta(u,v))$ given any queried location $\vx$ through a single forward pass and obtain the magnitude of the loss landscape $L(\vtheta,\vz)$.

\begin{figure}[t]
	\centering
	\includegraphics[width=0.98\linewidth]{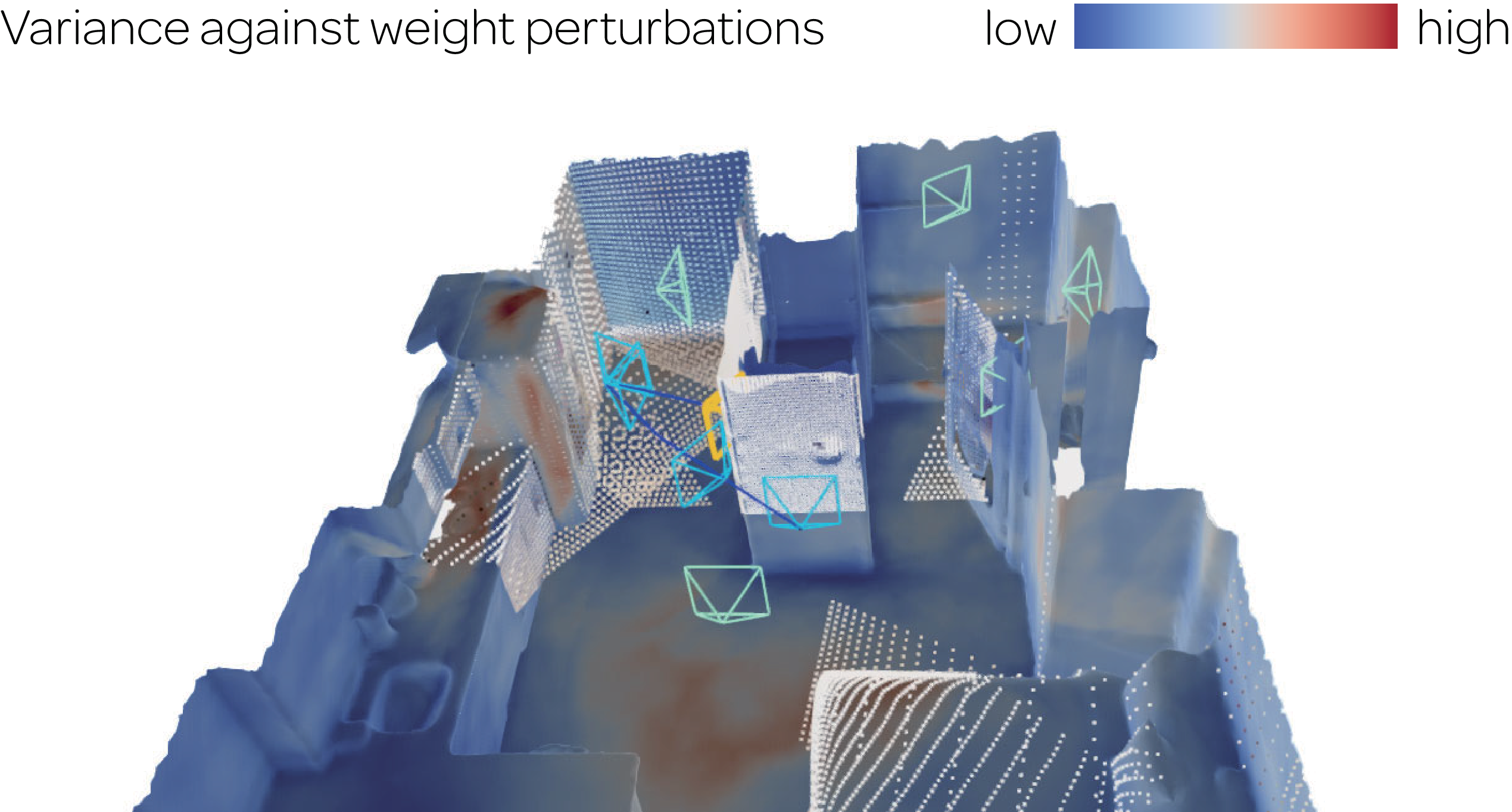}
	\caption{The functionality changes due to weight perturbations given the ground truth surface points $\vx \in \gD$. It can be noted that most high-variance regions (with warm colors) locate near the boundaries of space between explored (colored point cloud) and unexplored areas. }
	\label{fig:variability}
\end{figure}

We randomly pick a true surface point $\vx^+$ observed at $t=1$ and a false-positive zero-crossing point\footnote{A free-space point whose instant prediction $f(\vx^-;\vtheta^{200})\approx 0$} $\vx^-$ generated at~$t=200$ due to the continuous nature of the neural map. The prediction over the entire weight space is then calculated through forward passes given $\vtheta(u,v)$. As presented in~\cref{fig:loss_landscape}, by projecting the high-dimensional weight space onto the hyperplane, we can easily visualize the loss changes along the optimization path. Empirically, we observe evidently-different geometries for the true surface point and the false-positive one: the loss of the true surface point will be constrained in a low-loss basin, while the loss of the false-positive one stays along a sharp ridge that once jumps over a high-loss ridge into the valley at $t=200$ and then keeps ascending. 

The reason behind this phenomenon is straightforward. During continual learning, the parameters of the neural map undergo constant changes. The functionality of $f(\vx;\vtheta^t)$ will only remain stable in previously-observed areas with constant self-supervision (as verified in~\cite{Yan2021iccv, Sucar2021iccv, Ortiz2022rss} through a simple experience replay strategy). In not-well-explored areas, the functionality can easily change due to a lack of constraints. That is to say, the neural map is more susceptible to areas where the functionality changes the most against parameter perturbations:
\begin{equation}
	\vx = \arg\max \mathbb{V}_{\hat{\vtheta}\sim\mathit{N}(\vtheta,b^2\mathit{I})}[f(\mathbf{x};\hat{\vtheta})].
	\label{eq:variability}
\end{equation}

\begin{figure*}[t]
	\centering
	\includegraphics[width=0.99\linewidth]{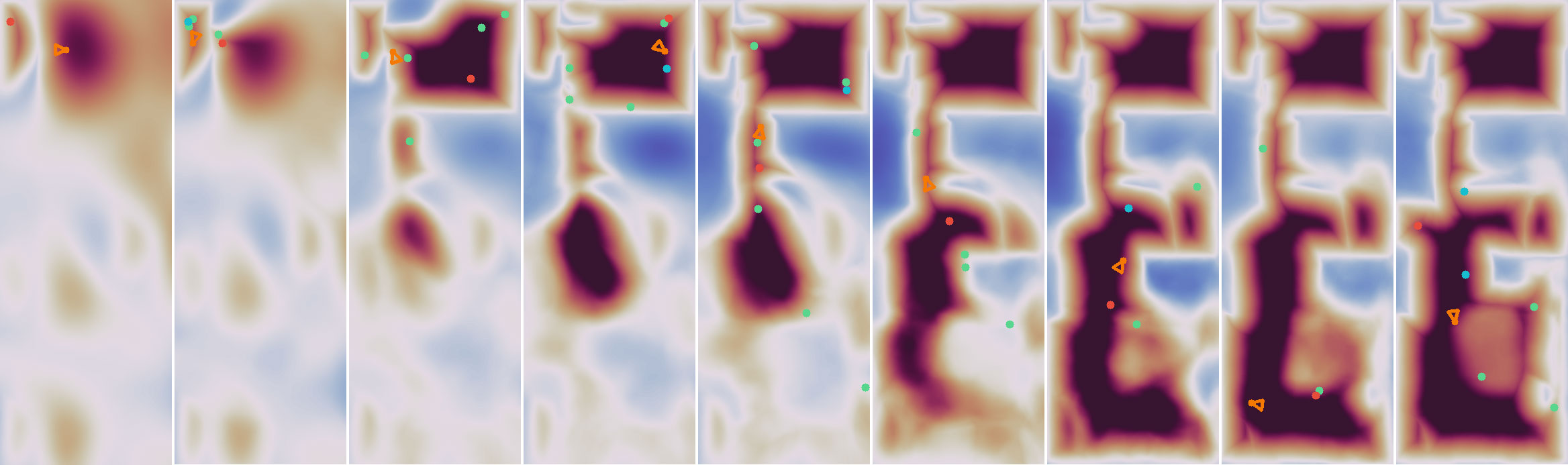}
	\caption{The evolution of the learned signed distance field through active neural mapping in 1000 steps. The proposed system is conducted in a receding horizon fashion. The target locations (green dots) are constantly pushed to the not-well-explored or not-well-trained regions for reaching a local equilibrium point. See the supplementary video for better visualization.}
	\label{fig:evolution}
\end{figure*}

The term in~\cref{eq:variability} is referred to as the \emph{artificial neural variability}~\cite{Xie2021anv} that shares similar concepts with the \emph{neural variability} in neuroscience~\cite{Orban2016neuron, Ma2006NN, Festa2021neuronal}: neuronal activity fluctuates over time given the same inputs, indicating the uncertainty of perceptual inference. By evaluating the prediction variability given points on the zero-crossing surfaces through weight perturbation, the false-positive ones and the true-positive ones can be evidently distinguished due to variable behaviors, and observations around the false-positive ones indicate high generalization cost (the second term of~\cref{eq:tradeoff}) as they land in sharp and unstable minima.

As illustrated in~\cref{fig:variability}, the functional sensitivity of~\cref{eq:variability} is directly linked with the data distribution of past observations that supervise the neural map, and explicitly indicates the prediction quality and uncertainty. High-variance regions are usually around the boundaries of space between explored and unexplored areas (where false-positive zero-crossing surfaces are generated). This is in common with the prevalent concept of the frontier. The differences lie in that the high-variance regions are naturally indicated by the neural map in a data-driven fashion instead of the heuristic design. Besides, unlike frontiers that rely mainly on adjacent occupancy status, regions with scarce data or with thin structures may also fall into a high-variance region in our case as they struggle to converge. Hence, all areas that are not accurately represented are taken into account. We refer readers to the supplementary video for a better understanding.

\subsection{An online active neural mapping system}
\label{subsec:system}
In this paper, we target a continuous signed distance function (SDF) of $f(\vx;\vtheta): \sR^3 \to \sR$ as the scene geometry representation, where the neural map $\vtheta$ is continually optimized and guides the mobile agent to not-well-explored areas. The system is implemented as four steps: 1) identifying the target viewpoints; 2) selecting the best target viewpoint; 3) navigating to the target location; 4) and optimizing the neural map parameters given newly-captured data.

The target view identification serves as finding the most distribution shift $\delta\vz^*$. A zero-mean Gaussian perturbation around the instant map parameters $\vtheta^t$ is performed every time a keyframe is selected, where the variance $\mathit{I}$ is set as the norm of recent parameter changes $|\vtheta^t-\vtheta^{t-1}|$. We sample points on the predicted zero-crossing surface to distinguish between the real surface points and the false-positive ones. This is close in spirit to the frontier-based method in a sample-based fashion. In practice, the top 10\% points with the highest variance $\mathbb{V}_{\hat{\vtheta}\sim\mathit{N}(\vtheta,b^2\mathit{I})}[f(\mathbf{x};\hat{\vtheta})]$ are selected and then clustered based on the geometrical similarity. To make the selected samples in sight, we place the target locations (green dots in~\cref{fig:evolution}) at a fixed distance along the surface normal $\nabla f(\vx;\vtheta^t)$, where the continuous and differentiable neural representation allows for convenient gradient computation through auto-differentiation. To determine the best view (the red dot in~\cref{fig:evolution}) among the target location candidates, we evaluate each cluster with three criteria: the maximum variance against parameter perturbations, the number of points within the cluster, and the distance between the cluster center and the current agent position. As illustrated in~\cref{fig:evolution}, the red dot and cyan dot are selected based on different criteria. Besides, a local planning horizon~\cite{Cao2021rss, Zhou2021ral, Batinovic2021ral} is adopted that prioritizes the target viewpoint candidates within the frustum bounding box. Therefore, the agent (the orange arrow in~\cref{fig:evolution}) will choose the best candidate in sight as the target view. 

Within each receding horizon loop $[t, t + k]$, the point-goal navigation for deciding the agent actions $\va^{t:t+k}$ and the continual learning for updating the map parameters $\vtheta^{t:t+k}$ are exactly the optimization process for maximizing generalization and minimizing forgetting, indicating the dynamics of $\va$ and $\vtheta$ to reach the equilibrium point of $\{\delta \vz^*, \vtheta^*\}$ within a local horizon. For point-goal navigation, we adopt the reinforcement-learning-based DD-PPO~\cite{Wijmans2019iclr} to reach the next target viewpoint. For incrementally updating the neural map, we adopt the experience-replay-based strategy of iSDF~\cite{Ortiz2022rss} with similar architecture and loss functions. It should be noted that other planner~\cite{Adamkiewicz2022ral, Kurenkov2022ral, Lee2022ral} and continual learning strategies~\cite{Yan2021iccv, Sucar2021iccv} can be naturally incorporated as optimizers that decide the optimization path to reach the local equilibrium point of $\delta^*$ and $\vtheta^*$.

%% file: srcs/evaluation.tex
\section{Experiments}
\label{sec:experiments}
Central to the paper is a novel target view identification module through weight perturbations and an online active mapping system with a 3D implicit neural representation. In this section, we evaluate the performance of the system through comprehensive experiments.

\subsection{Experimental Setup}
The experiments are conducted on a desktop PC with an Intel Core i7-8700 (12 cores @ 3.2 GHz), 32GB of RAM, and a single NVIDIA GeForce RTX 2080Ti.

\noindent\textbf{Data acquisition.} Our algorithm is conducted with the Habitat simulator~\cite{Habitat19iccv} and evaluated on the visually-realistic Gibson~\cite{Gibson2018cvpr} and Matterport3D datasets~\cite{Matterport2017_3DV}. The experiments are conducted in 1000/2000 steps depending on the scene scale.\footnote{A more thorough introduction of the test scenes and per-scene analysis are provided in the supplementary material.}~The system takes posed depth images at the resolution of $256\times256$ as inputs and outputs discrete action at each step. The action space consists of \texttt{MOVE\_FORWARD} by $6.5$$cm$, \texttt{TURN\_LEFT} and \texttt{TURN\_RIGHT} by $10^{\circ}$, and \texttt{STOP}. The mobile agent is randomly initialized in the traversable space at the height of $1.25$m. The field of view (FOV) is set to $90^{\circ}$ vertically and horizontally.

\noindent\textbf{Neural map architecture.} Our neural map is a single multi-layer perceptron (MLP) with 4 hidden layers and 256 units per layer. Following~\cite{Ortiz2022rss}, a softplus activation and a positional embedding are adopted, where the positional embedding is concatenated to the third layer of the network. The neural map is optimized using the Adam optimizer with a learning rate of $0.0013$.

\subsection{Evaluation metrics} 
We adopt the following metrics for evaluating the incrementally-updated neural map:

\noindent\textbf{MAD}~($cm$).~The mean absolute distance is evaluated by estimating the distance prediction through a forward pass on all vertices from the ground truth mesh. This metric defines the accuracy of the learned 3D neural distance field.


\noindent\textbf{FPR}~(\%).~The false-positive rate is calculated as the percentage of samples from the reconstructed mesh whose nearest distance to the ground truth mesh exceeds $5$$cm$. This metric defines the quality of the mesh extracted from the 3D continuous neural map.

\noindent\textbf{Comp.}. The completeness metrics are calculated from the ground truth vertices to the entire observations that are actively captured. By estimating per-vertex nearest distance to the past observations $\vz^{1:t}$, the percentage of points whose nearest distance is within 5cm (Comp. (\%)) and the mean nearest distance (Comp. ($cm$)) can be calculated to measure the active exploration coverage in 3D space.

%

\subsection{Comparisons against other methods}
\begin{table}
	\centering
	\caption{The coverage of the actively-captured data. See supplementary material for results on each scene for details.}
	\label{tab:baselines}
	\begin{tabular}{lcccc} 
		\toprule
		& \multicolumn{2}{c}{\emph{Gibson}} & \multicolumn{2}{c}{\emph{MP3D}} \\
		\cmidrule(lr){2-3} \cmidrule(lr){4-5}
		& \textbf{Comp. ↑} & \textbf{Comp. ↓} & \textbf{Comp. ↑} & \textbf{Comp. ↓}\\
		& (\%) & ($cm$) & (\%) & ($cm$) \\
		\toprule
		\textbf{Random}  & 45.80 & 34.48  &45.67 & 26.53 \\
		\cmidrule(lr){1-5}
		\textbf{FBE} &  68.91 & 14.42 & 71.18  & 9.78 \\
		\cmidrule(lr){1-5}
		\textbf{UPEN} & 63.30 & 21.09  & 69.06 &  10.60 \\
		\cmidrule(lr){1-5}
		\textbf{OccAnt}  & 61.88 & 23.25 & 71.72 & 9.40 \\
		\cmidrule(lr){1-5}
		\textbf{Ours} & \textbf{80.45} & \textbf{7.44}  & \textbf{73.15} & \textbf{9.11} \\
		\bottomrule
	\end{tabular}
	
\end{table}
We compare the proposed method against three relevant methods: FBE~\cite{Yamauchi1997frontier} aims to push the boundaries between unknown and known space for exploration;
OccAnt~\cite{Ramakrishnan2020eccv} anticipates the occupancy status in unseen areas and rewards the agent with accurate anticipation;
UPEN~\cite{Georgakis2022icra} tries to select the most uncertain path via the ensemble of occupancy prediction models. As the three methods utilize the 2D grid-based map representation, we evaluate the completeness of the actively-captured observations along the trajectory using the Comp.~(\%) and the Comp.~($cm$) metrics.

As presented in~\cref{tab:baselines}, the proposed active mapping system consistently outperforms the three competitors. It should be noted that FBE and UPEN adopt the same DD-PPO planner as ours for target goal navigation. Therefore, the efficacy of the proposed goal location identification strategy can be fairly evaluated. FBE relies purely on the voxel-based geometric information for identifying the frontiers, whereas the selection mechanism is manually designed that can easily ignore areas that have been explored with insufficient data. In contrast, the proposed method well quantifies the map uncertainty to achieve better performance. In terms of OccAnt, the agent occasionally moves back and forth as the goal location identification is trained through a rewarded mechanism, while the proposed goal location identification strategy and the local planning horizon guarantee stable exploration routes. UPEN adopts a deep ensemble based manner~\cite{deep_ensemble} to quantify the prediction uncertainty, which shares a similar idea with the proposed method regarding epistemic uncertainty reasoning. Nevertheless, UPEN simply generates multiple traversable path candidates towards a pre-defined unreachable goal location with a global RRT planner, where the map uncertainty merely ranks the path candidates to achieve the best information gain, while the proposed method better explores the environment by explicitly exploiting the neural variability for goal location identification and selection.

\subsection{Ablation study and system performance}
\begin{table}
	\centering
	\caption{Ablation study of the map quality regarding the SDF prediction (MAD), the reconstructed mesh (FPR), and the observation completeness (Comp.).}
	\label{tab:ablation}
	\begin{tabular}{llccc} 
		\toprule
		&& \textbf{MAD ↓} & \textbf{FPR ↓} & \textbf{Comp. ↑}\\
		&& ($cm$) & (\%) & (\%) \\
		\toprule	
		&\textbf{Random}      & 8.49  & 36.57  & 45.80  \\
		\cmidrule(lr){2-5}
		\multirow{4}{*}{\begin{sideways}\emph{Gibson}\end{sideways}} 
		&\textbf{Module 1} & 5.68  & 34.65  & 79.48\\
		\cmidrule(lr){2-5}
		&\textbf{Module 3} & 5.44 &  \textbf{25.09} & 76.19  \\
		\cmidrule(lr){2-5}
		&\textbf{Module 4} & 6.11  & 32.86 & 73.70 \\
		\cmidrule(lr){2-5}
		&\textbf{Ours} & \textbf{5.10} & 28.04 & \textbf{80.45}\\
		\midrule
		&\textbf{Random}      & 8.87  & 51.88 & 45.67  \\
		\cmidrule(lr){2-5}
		\multirow{4}{*}{\begin{sideways}\emph{MP3D}\end{sideways}} 
		&\textbf{Module 1} & 4.65 & 43.03 & 71.41 \\
		\cmidrule(lr){2-5}
		&\textbf{Module 3} & 6.06 & \textbf{39.05} & \textbf{74.63}  \\
		\cmidrule(lr){2-5}
		&\textbf{Module 4} & 4.30 & 47.99 & 67.75 \\
		\cmidrule(lr){2-5}
		&\textbf{Ours} & \textbf{4.29} &40.07  & 73.15 \\
		\bottomrule
	\end{tabular}
\end{table}

As mentioned in~\cref{subsec:system}, the proposed active neural mapping system allows drop-in substitutes to replace the existing modules. We provide detailed ablation studies to justify the reasonable design of each module.

\noindent\textbf{Module~1}: target view identification.~We replace the proposed weight perturbation module with MC-Dropout~\cite{Gal2016icml} (p=0.05) and a BALD~\cite{BALD} score to quantify the prediction uncertainty. The output is sampled five times in our experiments. As demonstrated in~\cref{tab:ablation}, the proposed goal location identification strategy achieves better results compared to the substitute. Although the uncertainty quantification method leads to comparable exploration efficiency, the involvement of Dropout layers leads to noisy and coarse geometry and inefficient inference.

\begin{figure}[t]
	\centering
	\includegraphics[width=0.98\linewidth]{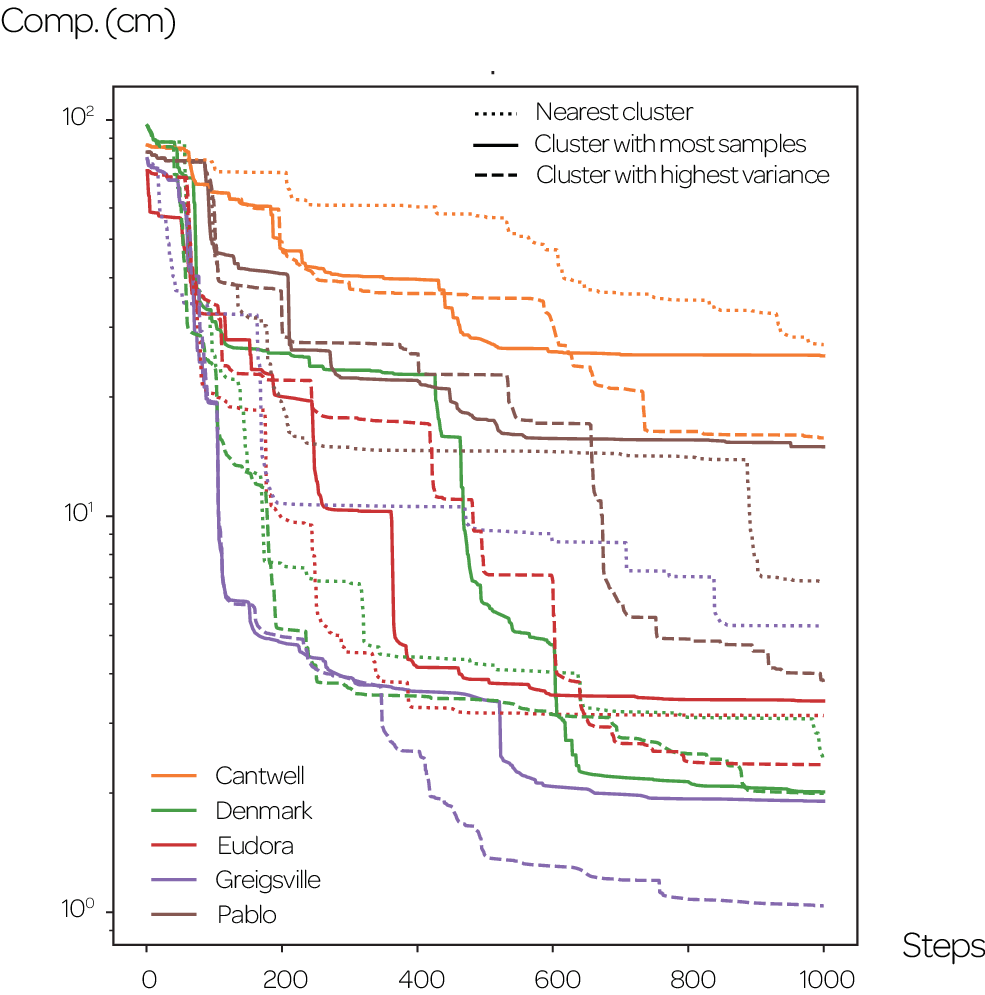}
	\caption{The effect of different candidate selection criteria (Module 2) on the Comp. ($cm$) ↓ metric in 1000 steps. Best viewed in color to see results on different scenes.}
	\label{fig:criteria}
\end{figure}

\noindent\textbf{Module~2}: best candidate selection. As illustrated in~\cref{fig:criteria}, we evaluate the performance of the three different selection criteria mentioned in~\cref{subsec:system}. Results on different scenes share a similar conclusion: selecting the highest variance regions will lead to the best performance. This result meets the arguments in~\cref{sec:pre} and~\ref{sec:method} to obtain the equilibrium point by maximizing generalization, or in other words, moving to the highest variance areas as~\cref{eq:variability}.

\begin{figure}
	\centering
	\includegraphics[width=0.98\linewidth]{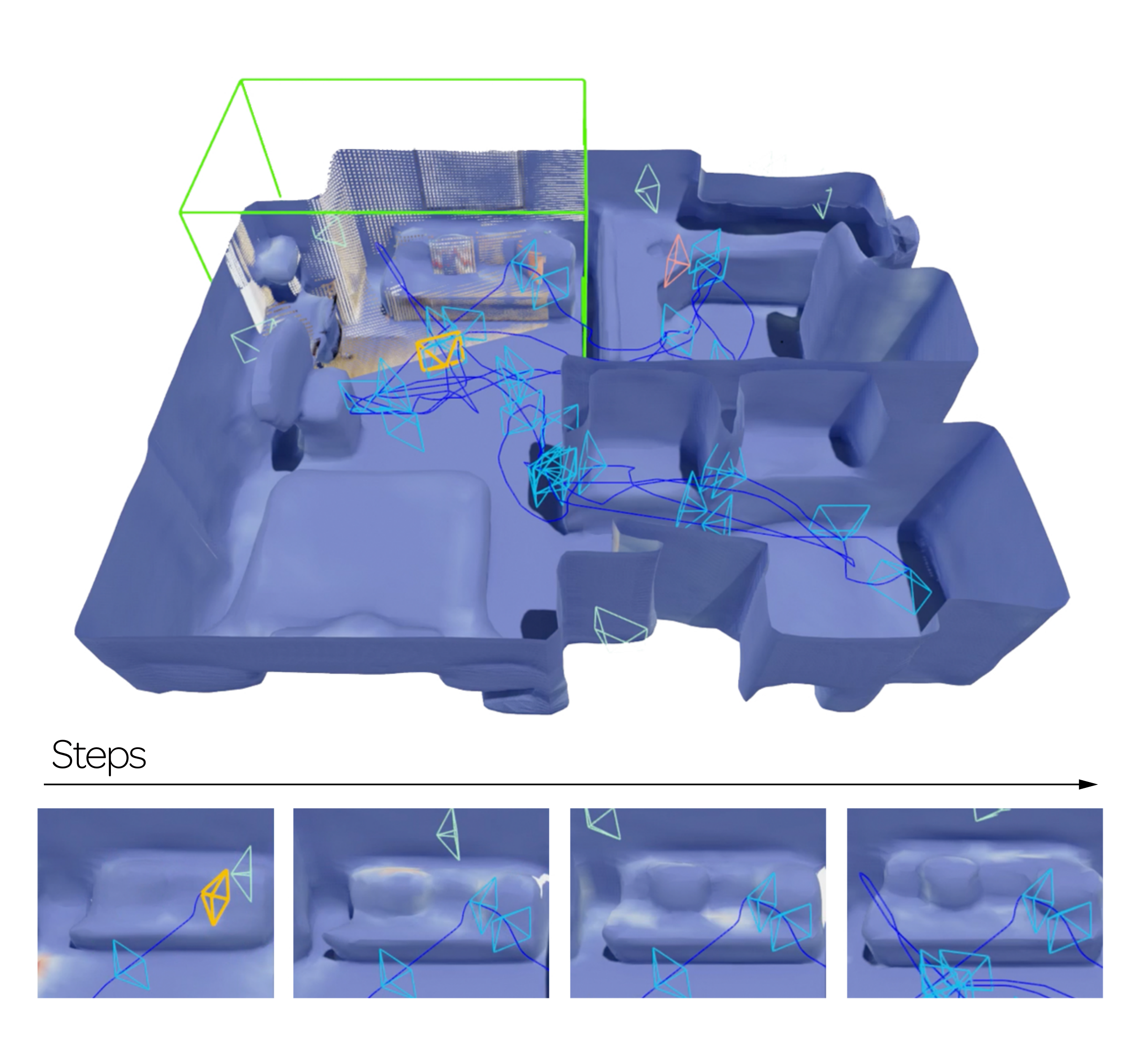}
	\caption{Continual learning of the scene geometry.}
	\label{fig:continual_learning}
\end{figure}

\begin{figure*}
	\centering
	\includegraphics[width=0.99\linewidth]{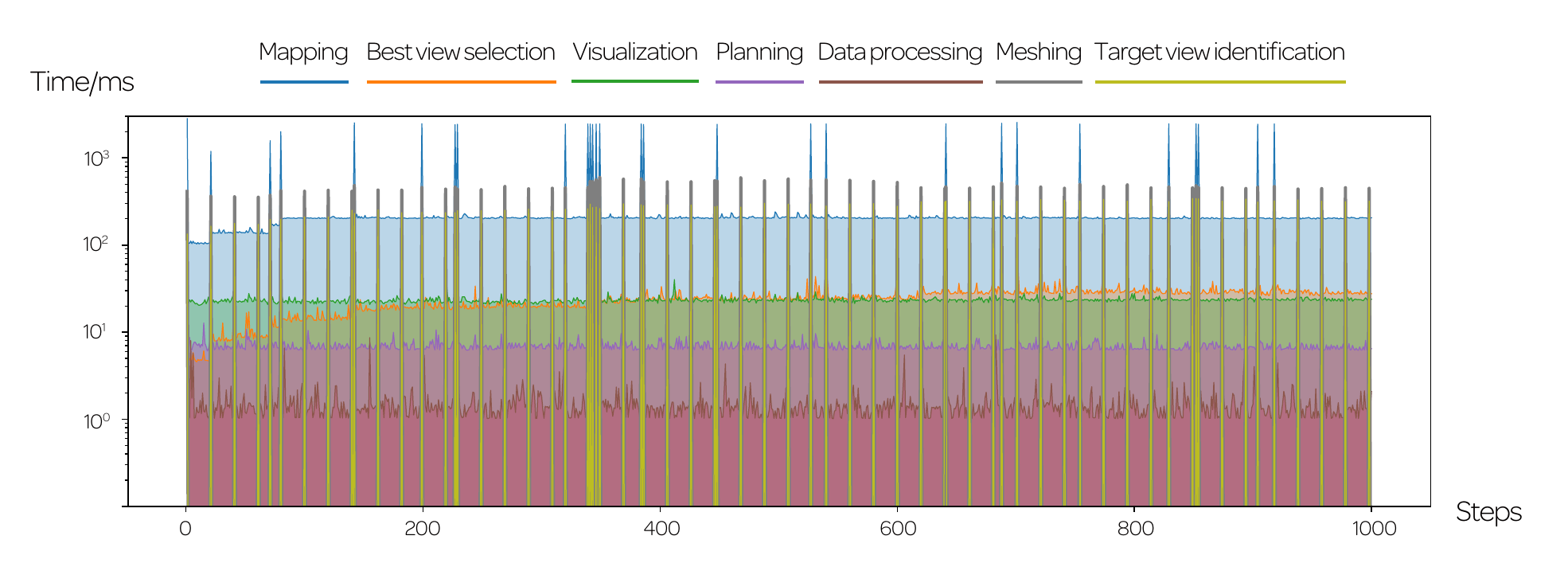}
	\caption{The runtime for each module. The impulse of the runtime is caused by a keyframe-based or windowed execution.}
	\label{fig:time}
\end{figure*}

\noindent\textbf{Module~3}: local planner. In our final setting (Ours) and for evaluating FBE/UPEN, we choose the DD-PPO$^+$ model trained on Gibson4+ and Matterport3D (train/val/test) for evaluating the Gibson validation sequences, and choose the DD-PPO$^\star$ model trained on Gibson2+ for evaluating the Matterport3D test sets to avoid the over-fitting issue. We here alter the model choice to further evaluate the performance on Gibson with the DD-PPO$^\star$ and on Matterport3D with DD-PPO$^+$.~Without pretrained data from Matterport3D,~DD-PPO$^\star$ results in degradation on large scenes and improvement in small ones, while DD-PPO$^+$ leads to more robust and balanced results.  We refer readers to the supplementary material for a more detailed per-scene analysis. The results further verify the efficacy of the proposed goal location identification strategy: a more powerful planner will bring better exploration results only if the goal location is properly decided.

\noindent\textbf{Module~4}: learning of the neural map. Different network architectures affect the convergence rate and the generation of false-positive zero-crossing surfaces. We further evaluate the active neural mapping system with a different network architecture: a single MLP with positional encoding~\cite{Mildenhall2020eccv} and ReLU activations. As the substitute architecture converges slower for high-frequency components~\cite{Tancik2020nips}, the reconstruction accuracy deteriorates compared to our final setting (Ours). Meanwhile, the exploration is slightly affected as the prediction in visited areas may still be inaccurate. Nevertheless, the system still works effectively given a different network architecture, suggesting the applicability to embracing the latest advances in implicit neural representations. 

\noindent\textbf{System performance.} In general, our system achieves promising reconstruction accuracy and completeness given limited steps. The computational cost for each module is illustrated in~\cref{fig:time}, where the runtime per step is 0.33s on average. The system is real-time capable and can be accelerated by reducing the per-frame iteration during continual learning. As illustrated in~\cref{fig:continual_learning}, the prediction of scene geometry over previously seen areas is continually improved during exploration, and the coverage of space continues to grow. The continual learning fashion allows for constant map optimization and lifelong learning of the scene.